\pdfoutput=1

\documentclass[11pt]{article}

\usepackage[final]{acl}

\usepackage{amsmath}
\usepackage{times}
\usepackage{latexsym}
\usepackage[normalem]{ulem}
\usepackage[T1]{fontenc}

\usepackage[utf8]{inputenc}

\usepackage{microtype}

\usepackage{inconsolata}

\usepackage{graphicx}

\usepackage{multirow} 
\usepackage{booktabs}       
\usepackage{amsmath}        
\usepackage{amsfonts}       

\usepackage{todonotes}

\title{\texttt{FedReFT}: Federated Representation Fine-Tuning with All-But-Me Aggregation}
\author{
 \textbf{Fatema Siddika\textsuperscript{1}\thanks{Equal contribution}}, 
 \textbf{Md Anwar Hossen\textsuperscript{1}\footnotemark[1]}, 
 \textbf{Juan Pablo Munoz\textsuperscript{2}},  \\
 \textbf{Tanya Roosta\textsuperscript{3}}
 \textbf{Anuj Sharma\textsuperscript{1}}, 
 \textbf{Ali Jannesari\textsuperscript{1}} 
 \\
 \textsuperscript{1}Iowa State University, Ames, USA \\
 \textsuperscript{2}Maro Systems, USA \\
 \textsuperscript{3}University of California, Berkeley
 \\
 \texttt{\{fatemask, manwar, anujs, jannesar\}@iastate.edu} \\
 \texttt{pablo.munoz@maro-systems.com}, tanya.roosta@gmail.com
}

\begin{document}
\maketitle
\begin{abstract}
Parameter-efficient fine-tuning (PEFT) adapts large pre-trained models by updating only a small subset of parameters. Recently, Representation Fine-Tuning (ReFT) has emerged as an effective alternative. ReFT shifts the fine-tuning paradigm from updating model weights to directly manipulating hidden representations that capture rich semantic information, and outperform state-of-the-art PEFTs in standalone settings. However, its application in Federated Learning (FL) remains challenging due to heterogeneity in clients' data distributions, model capacities, and computational resources. To address these challenges, we introduce \textbf{Fed}erated \textbf{Re}presentation \textbf{F}ine-\textbf{T}uning (FedReFT), a novel approach to fine-tune clients' hidden representations. FedReFT applies sparse intervention layers to steer hidden representations directly, offering a lightweight and semantically rich fine-tuning alternative ideal for edge devices. However, representation-level updates are especially vulnerable to aggregation mismatch under different task heterogeneity, where naive averaging can corrupt semantic alignment. To mitigate this issue, we propose All-But-Me (ABM) aggregation, where each client receives the aggregated updates of others and partially incorporates them, enabling stable and personalized learning by balancing local focus with global knowledge. We further design an adaptive update strategy inspired by Test-Time Computing (TTC) to balance local and global contributions under heterogeneous conditions. FedReFT achieves state-of-the-art performance on commonsense reasoning, arithmetic reasoning, and GLUE benchmarks, while delivering $1\times $–$49\times$ higher parameter efficiency compared to leading LoRA-based methods. The paper code is available at
 \href{https://anonymous.4open.science/r/FedReFT}{Anonymous Repository}
\end{abstract}

\section{Introduction}
\label{sec:introduction}
Fine-tuning has emerged as a core strategy for adapting large language models (LLMs) to various downstream tasks, allowing for a broad generalization from minimal task-specific data \cite{ding2023parameter,ziegler2019fine}.  However, traditional full fine-tuning is computationally expensive and memory-intensive, which poses scalability challenges. This is further amplified in resource-constrained environments, such as smartphones, where full model updates are often infeasible due to limited resources. 
To address these challenges, parameter-efficient fine-tuning (PEFT) methods, such as Adapter Tuning \cite{houlsby2019parameter}, BitFit \cite{zaken2022bitfit}, Prefix Tuning \cite{li2021prefix}, Prompt Tuning \cite{lester2021power}, and Low-Rank Adaptation (LoRA) \cite{Hu21}, have been proposed.  These methods significantly reduce the cost of adaptation by updating only a small subset of model weights. 

PEFT has emerged as the preferred method for efficiently adapting large language models (LLMs) without sacrificing performance. However, most PEFT approaches assume centralized data access, which is unrealistic in many real-world scenarios where data is distributed across users or devices with varying tasks and privacy concerns. Federated Learning (FL) offers a solution by enabling collaborative model training without centralizing data, but prior FL work often emphasizes task-specific tuning rather than learning generalizable representations. In practice, clients frequently work on diverse or specialized tasks, making global representation learning both more difficult and more essential.

While PEFT typically modifies model weights, recent interpretability research highlights the potential of hidden representations, which encode rich semantic information. ReFT~\cite{Wu2024} leverages this by training small interventions that act as lightweight transformations on the model’s internal representations, steering model behavior for downstream tasks without altering the original weights. This representation-level adaptation enables ReFT to achieve stronger performance than weight-based methods such as LoRA. Despite ReFT’s success in centralized settings, it has yet to be adapted to FL setting, where challenges such as data heterogeneity, varying model capacities, and limited computational resources complicate aggregation and reduce effectiveness. To investigate the challenges of representation-level fine-tuning in heterogeneous federated settings and assess the effectiveness of our proposed aggregation strategy, we formulate the following research questions: \\
\textbf{RQ1:} How can representation-level updates be aggregated in FL while preserving semantic alignment across task-heterogeneous clients?  \\
\textbf{RQ2:} Is simple weighted averaging sufficient for aligning semantically rich, hidden representations, or is a more robust and personalized strategy required to maintain local semantics while leveraging global knowledge?

To address these challenges, we propose \textbf{Fed}erated \textbf{Re}presentation \textbf{F}ine-\textbf{T}uning (FedReFT), a framework for personalized and parameter-efficient federated fine-tuning. FedReFT extends ReFT by injecting lightweight intervention components (sparse low-rank matrices $W$, $R$, $b$) directly into hidden representations, making it suitable for resource-limited edge devices. To avoid semantic misalignment from naive aggregation such as FedAvg \cite{mcmahan2017communication}, we introduce \textit{All-But-Me (ABM)} aggregation, which builds a global aggregated intervention by computing the geometric median over updates from all other clients.
The key contributions of our work are as follows:\\
\textbf{Contribution 1:} We address a critical gap in adapting ReFT to federated learning by introducing FedReFT, the first framework that enables personalized and parameter-efficient fine-tuning through sparse representation-level interventions, making it effective for resource-constrained clients.\\  
\textbf{Contribution 2:} To support this framework, we propose the \textit{All-But-Me (ABM)} aggregation strategy, specifically designed for representation-level interventions. ABM mitigates semantic misalignment caused by naive averaging and enables stable, personalized collaboration under heterogeneous conditions. Furthermore, FedReFT incorporates an adaptive mixing mechanism inspired by \textit{Test-Time Computing (TTC)} to dynamically learn how to combine local and ABM intervention parameters for each client.\\
\textbf{Contribution 3}: We evaluate the framework by simulating task heterogeneity, i.e., assigning different tasks to clients, all derived from a common dataset. This setup mimics real-world scenarios where clients pursue distinct objectives over structurally similar data, allowing us to evaluate the effectiveness of FedReFT and ABM under realistic conditions.
The rest of the paper is organized as follows. Section~\ref{sec:problem} defines the problem and challenges in heterogeneous FL. Section~\ref{sec:methodology} presents our FedReFT framework and ABM aggregation. Section~\ref{sec:experiments} reports experimental results, and Section~\ref{sec:conclusion} concludes with insights and future directions. Additional details are provided in the appendix.  
\begin{figure}[htbp]
    \centering
    \includegraphics[width=0.99\linewidth]{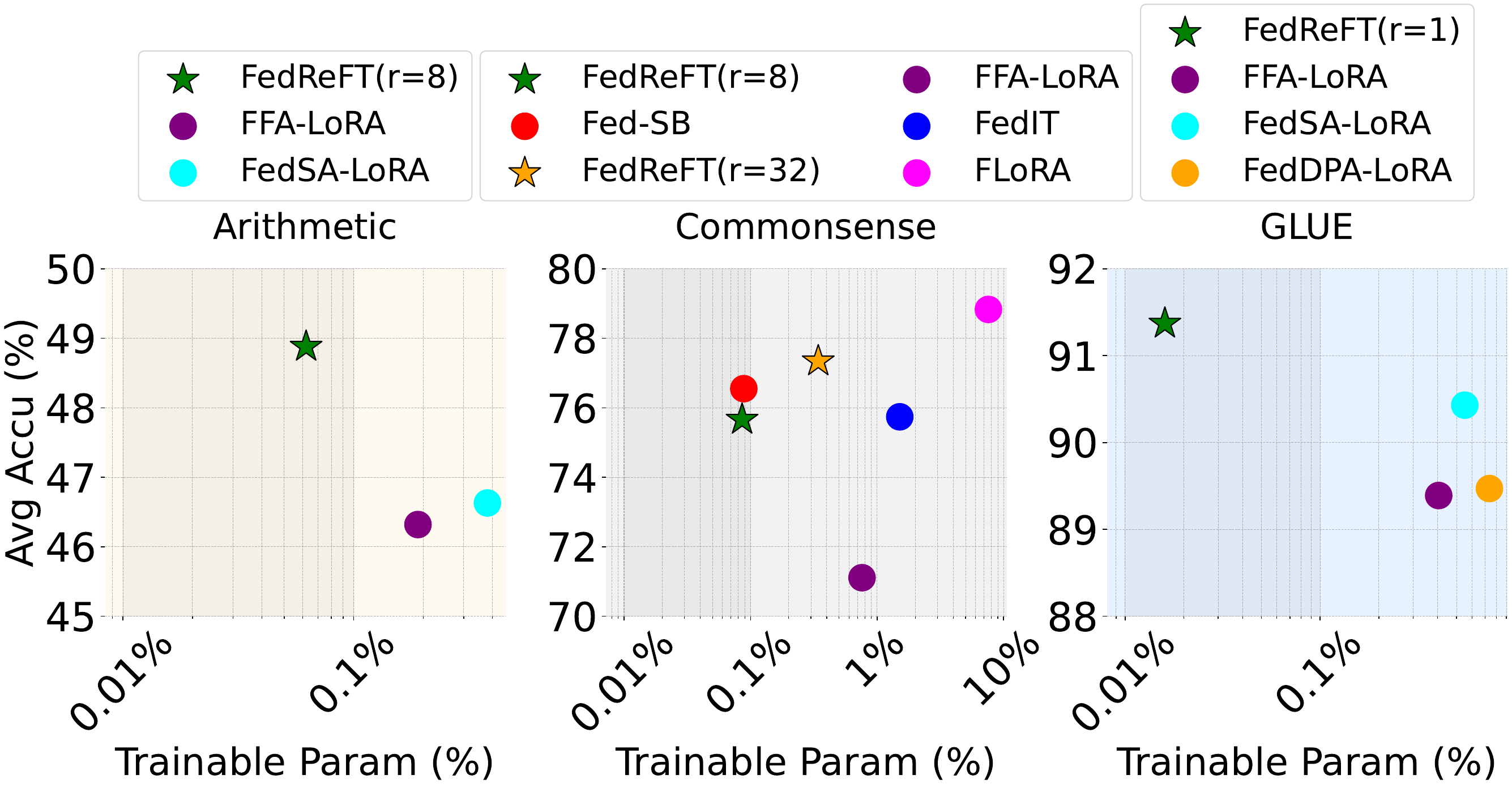}
    \caption{Average accuracy vs. trainable parameters (\%) for federated PEFT methods on Arithmetic, Commonsense, and GLUE benchmarks using LLaMA-3 8B, LLaMA-3.2B, and RoBERTa-large models, respectively. FedReFT attains state-of-the-art accuracy while training far fewer parameters, improving communication efficiency and reducing transmission cost in FL.}
    \label{fig:trainable_param}
\end{figure}
\section{Problem Formulation and Motivation}
\label{sec:problem}
In this section, we motivate applying ReFT to FL and formalize the problem of personalized representation adaptation. While ReFT enables parameter-efficient updates in the representation space, its use in FL is hindered by task heterogeneity, semantic misalignment, and unstable aggregation. Therefore, we define the objective as achieving parameter-efficient and semantically aligned adaptation in FL with ReFT.

\textbf{Challenge 1: LoReFT in FL Settings (RQ1)}: ReFT ~\cite{Wu2024} offers an attractive alternative by modifying hidden activations instead of model weights. By intervening directly in structured semantic subspaces, ReFT supports interpretable, modular, and task-aligned adaptation, particularly advantageous in task-heterogeneous FL settings. However, deploying full ReFT in federated settings can be challenging, as transmitting high-dimensional representation updates increases communication overhead. Moreover, when clients employ models with varying capacities or architectures, aligning representation spaces becomes nontrivial, making integration across clients difficult.


To bridge this gap, we adopt Low-Rank Linear Subspace ReFT (LoReFT)\cite{Wu2024}, which is a lightweight ReFT variant that constrains interventions to a learnable low-rank subspace. This design significantly reduces overhead while maintaining semantic control, making it a promising candidate for FL. We follow the LoReFT intervention formulation from \cite{Wu2024} on hidden representations $h \in \mathbb{R}^d$  which is defined as:
\begin{equation}
    \Phi_{\text{LoReFT}}(h) = h + R^\top (W h + b - R h),
    \label{eq:loreft}
\end{equation}
where, $W \in \mathbb{R}^{r \times d}$ is a low-rank projection matrix with 
$d$ as the representation dimension and $r$ as the subspace intervention dimension, $R\in \mathbb{R}^{r \times d}$ is a low-rank projection matrix with orthonormal rows, and $b \in \mathbb{R}^r$, with $r \ll d$. This structure, inspired by Distributed Interchange Intervention (DII)~\cite{geiger2024finding}, enables semantically grounded, low-rank adaptation suitable for scalable and privacy-preserving FL.
Despite its efficiency, applying LoReFT in FL raises several non-trivial challenges:
(i) LoReFT modifies internal representations that are sensitive to client-specific data distributions. Aggregating these interventions naïvely using FedAvg can cause semantic interference or collapse.
(ii) Without global synchronization, low-rank updates may evolve in divergent directions, especially when tasks are dissimilar.
(iii) Applying shared LoReFT interventions across clients risks overfitting to shared patterns while ignoring local semantics. 
Considering all these challenges, the major research question is:\\
\textit{Can representation-level adaptation via LoReFT achieve personalization and stability in federated environments without collapsing under task and data heterogeneity?}\\
FedReFT uses All-But-Me(ABM) aggregation to robustly combine intervention parameters while preserving personalization in heterogeneous FL.
\textbf{Challenge 2: Federated Fine-Tuning under Task Heterogeneity (RQ1 \& RQ2)}: 
A central motivation of our work is to address task heterogeneity in real-world FL, where clients perform fundamentally different tasks rather than optimizing a shared objective. For example, clients may work on
distinct reasoning tasks within natural language question answering that demand different semantic skills. While centralized fine-tuning has proven effective for such tasks, it assumes access to all data, which is unrealistic in decentralized settings. In FL, each client sees only a local, task-specific subset of the broader reasoning space, leading to highly heterogeneous training distributions, a common challenge in multi-department or cross-domain deployments.  This raises the question:
\textit{How can we learn a global representation that generalizes across tasks when each client trains only on a fragment of the broader task distribution?}
Standard methods like FedAvg~\cite{mcmahan2017communication} struggle in this regime, as they average semantically misaligned updates, often resulting in degraded performance or collapsed representations. Formally, let each client $i$ have a dataset $\mathcal{T}_i = \{X_i, Y_i\}$ and optimize a personalized model $\boldsymbol{\theta}_i$ by solving:
\begin{equation}
    \min_{\boldsymbol{\Theta}} \ \frac{1}{N} \sum_{i=1}^{N} \mathcal{L}(X_i, Y_i, \boldsymbol{\theta}_i),
\end{equation}
where $\mathcal{L}$ is the task-specific loss and $\boldsymbol{\Theta} = \{\boldsymbol{\theta}_i\}_{i=1}^{N}$ is the set of client-specific models.  Our proposed method, FedReFT, can successfully address this research challenge. FedReFT enables scalable, personalized representation learning across heterogeneous tasks, allowing global reasoning capabilities to emerge from decentralized, task-specific updates.
\textbf{Challenge 3: Learnable Parameter Sharing with the Server (RQ2)}: When applying ReFT \cite{Wu2024} in a FL setting from the perspective of learnable parameter sharing, a fundamental question is:\\
\emph{Which of these parameters should be communicated to the server for collaborative aggregation?}
\begin{figure*}[h]
  \centering\includegraphics[width=1\linewidth]{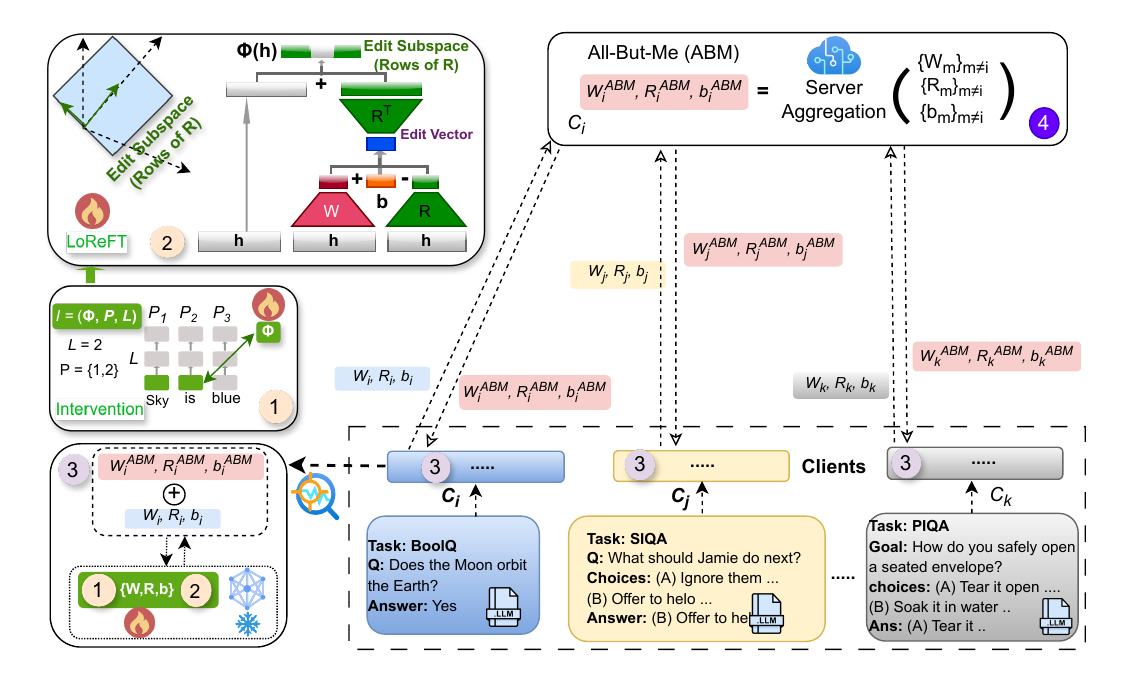}
     \caption{\textbf{FedReFT with ABM Aggregation.} Clients cross-task demonstrate personalization while maintaining alignment with the global representation.
(1)-(2): Each client applies LoReFT\cite{Wu2024} interventions to train learnable parameter \{$W$, $R$, $b$\} to modify hidden representations $h$ in a low-rank edit subspace. 
(3): Clients fine-tune \{$W$, $R$, $b$\} locally and partially fuse received \textit{All-But-Me} aggregated updates with their own. (4): The server performs ABM aggregation using the geometric median over other clients’ intervention parameters to generate $W_k^{\text{ABM}}, R_k^{\text{ABM}}, b_k^{\text{ABM}}$.} 
\label{fig:FedReF}
\end{figure*}
In FedReFT, each client fine-tunes hidden representations by introducing learnable low-rank intervention parameters $W$, $R$, and a bias $b$ into a frozen backbone model. Sharing only some part of the intervention parameters leads to incomplete information transfer and breaks the low-rank structure critical for generalization. 
$W$ projects representations into a low-dimensional space, and $R$ reconstructs them; omitting either disrupts compositionality and limits alignment across heterogeneous clients. In Table \ref{tab:ablation_uplink_strategy}, empirical results show that partial sharing significantly degrades performance and representation alignment under task heterogeneity. Therefore, FedReFT shares the full set of learnable intervention parameters $(W, R, B)$ from each client with the server. 
\section{Methodology}
\label{sec:methodology}
In this section, we introduce FedReFT, designed to address the challenges we discussed in the previous section. An illustrative overview of FedReFT is shown in Figure \ref{fig:FedReF}.
\subsection{Intervention Parameter Sharing Strategies}
An intervention is a small, learnable operation inserted into a model’s hidden layers that slightly modifies its internal representations, known as activations. Instead of updating the model’s weights, it transforms these hidden features using a few parameters in Equation \ref{eq:loreft}; a low-rank projection matrix $\mathit{R} \in \mathbb{R}^{r \times d}$ that captures compact feature subspaces, a transformation matrix $\mathit{W} \in \mathbb{R}^{r \times d}$ that maps these subspaces back into the model’s space, and a bias vector $\mathit{b} \in \mathbb{R}^{r}$ that shifts the representation to adjust how information flows through the network. We explore different strategies for sharing local intervention parameters with the server to reduce communication overhead while maintaining personalization. These strategies are summarized in Appendix \ref{abl:Intervention_Parameter_Sharing_Strategy} and represent different trade-offs between expressiveness and communication efficiency. After observing results and analysis, all the experiments shared the full Intervention parameters.
\subsection{Intervention Design for Federated Classification Tasks}
Following the formulation in ReFT~\cite{Wu2024}, for a given client, we define the classification head $H_\psi$ with parameters $\psi = \{W_o, b_o, W_d, b_d\}$ which operates on the CLS token representation $z \in \mathbb{R}^d$ from the final layer:
\begin{equation}
H_\psi(z) = \text{softmax}\left(W_o \cdot \tanh(W_d z + b_d) + b_o\right).
\end{equation}
We jointly optimize the intervention parameters $\phi$ and the classifier $\psi$ using cross-entropy loss over input $x$ and label $y$:
\begin{equation}
\min_{\phi, \psi} \left\{ -\log H_\psi\left(y \mid z_{\phi}(x)\right) \right\}.
\end{equation}
\subsection{All-But-Me (ABM) Aggregation on Server}
\label{sec:ABM}
In heterogeneous FL, the integration of shared knowledge without compromising local task-specific adaptation remains a core challenge. Standard aggregation methods, such as FedAvg~\cite{mcmahan2017communication}, which averages client models into a single global model, are often suboptimal in non i.i.d. scenarios. They risk overwriting valuable client-specific representations and rely on fixed mixing weights that may further reduce personalization.
To overcome these limitations, we propose the \textit{All-But-Me (ABM)} aggregation strategy. Instead of initializing clients with a global model, each client continues to update its local parameters while partially incorporating knowledge aggregated from other clients. Specifically, each client $k$ receives a robustly aggregated set of intervention parameters $\{\mathit{W}^{\text{ABM}}_k, \mathit{R}^{\text{ABM}}_k, \mathit{B}^{\text{ABM}}_k\}$, calculated from the updates of all other clients using a geometric median:
\begin{equation}
X=\{ \mathit{W}, \mathit{R}, \mathit{B} \}, \quad X_k^{\text{ABM}} = \text{ABM}( \left\{ X_m^{\text{c}} \right\}_{m \ne k}).
\end{equation}
\textbf{ABM via Geometric Median.}
The geometric median (also known as the spatial or $L_1$ median) offers a robust alternative to the arithmetic mean, particularly under client heterogeneity and adversarial conditions~\cite{maronna2006yohai, weiszfeld1937point}. Given a set of vectors $\mathcal{S} = \{x_1, \ldots, x_n\} \subset \mathbb{R}^d$, it is defined as:
\begin{equation}
x^* = \arg\min_{x \in \mathbb{R}^d} \sum_{i=1}^n \|x - x_i\|_2,
\end{equation}
which minimizes the sum of Euclidean distances to all elements in the set. This estimator is robust to outliers and misaligned updates, making it well-suited for federated settings.  We instantiate the ABM function using the geometric median, where each client $k$ receives an aggregated intervention vector computed from $\mathcal{S}_k = \{x_m\}_{m \ne k}$:
\begin{equation}
\text{ABM}(\mathcal{S}_k) = \arg\min_{x \in \mathbb{R}^d} \sum_{x_m \in \mathcal{S}_k} \|x - x_m\|_2.
\end{equation}
\begin{table}[!htbp]
\renewcommand{\arraystretch}{0.99}
\setlength{\tabcolsep}{2.7pt}
\centering
\caption{Comparison of FedAvg, Mean\_ABM, and GeoMedian\_ABM across commonsense, arithmetic, and GLUE tasks. Geometric median-based ABM aggregation achieves better accuracy over all other aggregation strategies.}
\label{tbl:GeoMed_results_different_task}
\begin{tabular}{l l c c}
\toprule
\textbf{Task, Model} & \textbf{Agg. Method} & \textbf{Acc} & \textbf{$\Delta$Acc$\uparrow$}  \\
\midrule

Commonsense & FedAvg         & 74.89 & \\
       LLaMA-2 7B                 & Mean\_ABM      & 75.15 & +0.25  \\
                        & \textbf{GeoMed\_ABM} & \textbf{76.55} & \textbf{+1.66}       \\
\midrule
Arithmetic  & FedAvg         & 24.87 &  \\
         LLaMA-2 7B               & Mean\_ABM      & 25.12 & +0.25  \\
                        & \textbf{GeoMed\_ABM} & \textbf{26.09} & \textbf{+1.22}      \\
\midrule
GLUE           & FedAvg         &  93.57 & \\ 
    RoBERTa                    & Mean\_ABM      &94.10 & +0.53 \\
                        & \textbf{GeoMed\_ABM} & \textbf{94.91} & \textbf{+1.34}      \\
\bottomrule
\end{tabular}
\end{table} 
To solve this optimization efficiently, we employ Weiszfeld’s algorithm, an iterative method known to converge under mild conditions. Details of the algorithm are provided in Table \ref{tbl:GeoMed_results_different_task}, Figure \ref{fig:ablation_aggregation_methods_comparison}, and Appendix \ref{abl:aggregation_methods}. By avoiding direct averaging and incorporating semantically meaningful low-rank intervention updates through robust aggregation, ABM enables each client to benefit from the knowledge of others without sacrificing local personalization. This approach enhances stability and generalization across non-i.i.d. and task-heterogeneous FL environments. Geometric median has complexity of O(T·d), with T being the number of iterations and d the number of parameters.


\subsection{Local Model Update with ABM Aggregation using Test-Time Computing}

In FedReFT, local interventions capture client-specific semantics, while ABM aggregation conveys shared global knowledge. However, naively averaging representation-level updates can risk disrupting semantic consistency across clients, potentially leading to misaligned feature spaces. Moreover, uniform aggregation overlooks client heterogeneity, limiting adaptability in diverse edge environments. 

These challenges highlight the need for a mechanism that can dynamically balance personalization and global alignment. Without such adaptation, clients may either overfit to their own data or lose semantic fidelity when forced into uniform global updates. FedReFT introduces an adaptive mixing strategy inspired by Test-Time Computing (TTC)~\cite{wang2021tent}, which learns client-specific coefficients to combine local and ABM aggregated intervention parameters. Unlike conventional fine-tuning that updates model weights, TTC performs lightweight optimization during inference, using a small, controlled compute budget to iteratively refine predictions and learn adaptive mixing parameters.
Here, unconstrained logits $\beta \in \mathbb{R}$ define mixing weights
$\alpha = \sigma(\beta) \in (0,1)$ via the sigmoid $\sigma(\cdot)$.
For any intervention tensor $\mathit{X} \in {\mathit{W}, \mathit{R}, \mathit{B}}$,
the mixed parameter is:
\begin{equation}
\mathit{X}_{\text{mixed}}(\alpha) \;=\; 
\alpha\,\mathit{X}^{\text{local}} + (1-\alpha)\,\mathit{X}^{\text{ABM}}.
\label{eq:mixed_update}
\end{equation}
TTC runs for $S$ steps over $B$ calibration batches of clients' test data.
At each step, it injects $\mathit{X}_{\text{mixed}}(\alpha)$ into the model
interventions, computes a forward pass to evaluate the test loss
$\mathcal{L}_{\text{test}}$, backpropagates gradients
$\nabla_\beta \mathcal{L}$ while keeping $\mathit{X}^{\text{local}}$ and
$\mathit{X}^{\text{ABM}}$ frozen, and then updates
$\beta \leftarrow \beta - \eta \nabla_\beta \mathcal{L}$.
The gradient
$\nabla_\beta \mathcal{L} = \nabla_\alpha \mathcal{L}\,\sigma(\beta)\big(1-\sigma(\beta)\big)$
directly indicates whether increasing $\alpha$ reduces the test loss:
negative gradients push $\beta$ higher (increasing $\alpha$), while
positive gradients push it lower. Over iterations, $\beta$ converges to:
\begin{equation}
\begin{aligned}
\beta^* \;=\; \arg\min_{\beta}\;
\mathcal{L}_{\text{test}}\!\big(\mathit{X}_{\text{mixed}}(\sigma(\beta))\big)
\;+\; \\ \lambda_1 \mathcal{H}(\alpha)
\;+\; \lambda_2 \mathcal{C}(\alpha)
\;+\; \lambda_3 \mathcal{D}(\alpha),
\end{aligned}
\label{eq:arg_min_lamda}
\end{equation}

where $\mathcal{H}$ (entropy) discourages collapse, $\mathcal{C}$ (consistency)
promotes stable mixing across keys, and $\mathcal{D}$ (diversity) avoids extreme
$\alpha$ saturation. \\
\textbf{Final Mixed Interventions.}
The optimal $\alpha^* = \sigma(\beta^*)$ is used to mix the local and ABM
intervention parameters:
\begin{equation}
\begin{aligned}
\mathit{X}_k^{\text{new}} &= \alpha^* \mathit{X}_k^{\text{local}}
+ (1 - \alpha^*) \mathit{X}_k^{\text{ABM}}, \\
\mathit{X} &\in \{\mathit{W}, \mathit{R}, \mathit{B}\}.
\end{aligned}
\end{equation}
Before the next local training with $\mathit{R}_k^{\text{new}}$, we apply an
orthogonal transformation to $\mathit{R}_k^{\text{new}}$ to preserve the original
property of $R$. We continue the discussion in Appendix \ref{abl:TTC_VS_Balanced_mixing} and 
Table \ref{tbl:baseline_first_commmonsense_ablation}, \ref{tbl:glue_task_ablation}, \ref{tbl:math_setup_1_2_resutls_ablation}, and \ref{tbl:math_setup_3_baseline_ablation} show that the performance of TTC-based adaptive mixing is over the balanced mixing. \\
\textbf{Computational Overhead.}
The TTC process costs $S \times B$ forward/backward passes per client.
With $N$ clients, $K$ intervention parameters, batch size $B$, $L$ layers, and
hidden width $d_h$, an approximate FLOPs is $\mathcal{O}\!\left(S \cdot M \cdot N \cdot K \cdot B \cdot L \cdot d_h^{2}\right)$.
Amortized over $T$ communication rounds and $|\mathcal{D}_{\text{test}}|$
inference batches, the effective per-inference overhead scales as
$\frac{S \cdot B}{T \cdot |\mathcal{D}_{\text{test}}|}$.
Here, small $S{=}50$ and $B{=}8$ keep the overhead modest while still improving
robustness under shift.
\begin{table*}[h!]
\setlength{\tabcolsep}{2.8pt}
\centering
\caption{Performance of LLaMA-3.2 3B across five commonsense reasoning tasks with Mixed-Task (MT) setup, where clients train on heterogeneous task mixtures to promote generalizable representations. \textsuperscript{*}Performance results of all baseline methods and the experimental setup are taken from \cite{singhal2025fed}. \textbf{Trainable Parameter (TP) Efficiency Rank 8} quantifies the relative parameter efficiency of FedReFT(R8), indicating how many times fewer trainable parameters it requires compared to baseline methods.}
\label{tab:fed-sb-performance}
\begin{tabular}{lccccccccc}
\toprule
\textbf{Method} & \textbf{R} & \textbf{TP(M) $\downarrow$} & \textbf{TP Effi. (R8)} $\downarrow$ & \textbf{BoolQ} & \textbf{PIQA} & \textbf{SIQA} & \textbf{HellaS.} & \textbf{WinoG} & \textbf{Avg Acc$\uparrow$}\\
\midrule
FedIT\textsuperscript{*}         & 32  & 48.63  & 17.68$\times$ & 62.99 & 81.50 & 73.13 & 76.83 & 71.51 & 75.74 \\
FFA-LoRA\textsuperscript{*}      & 32  & 24.31  & 8.84$\times$ & 62.87 & 80.03 & 68.53 & 70.02 & 65.56 & 71.11 \\
Fed-SB \textsuperscript{*}       & 120 & 2.83 & 1.03$\times$ &64.86 & 81.66 & 74.87 & 81.67 & 75.22 & 75.66 \\ 
Fed-SB \textsuperscript{*}       & 200 & 7.8 & 2.84$\times$ &66.66 & 83.79 & 77.22 & 85.42 & 79.56 & 78.53 \\ 
\midrule
\multirow{3}{*} {\textbf{FedReFT(ours)}}
 & 4 & 1.38 & 0.5$\times$ &  63.35 & 82.72 & 72.96 & 91.37 & 69.70 & 76.02 \\
  & 8 & 2.75 &  1.0$\times$ & 65.50 & 82.32 & 73.28 & 91.43 & 70.24 & 76.55\\

 & \textbf{16} & \textbf{5.5} &  \textbf{1.0$\times$} & \textbf{64.56}  & \textbf{82.20} & \textbf{75.95} & \textbf{89.80} &  \textbf{80.59}  & \textbf{78.62}\\
\bottomrule
\label{tbl:baseline_first_commmonsense}
\end{tabular}
\end{table*}

\begin{table*}[t]
\setlength{\tabcolsep}{2.4pt}
\caption{Performance of Federated fine-tuning of Llama-3.2 3B across eight commonsense reasoning datasets in a highly data-heterogeneous setting, which is denoted as Distinct Task (DT). \textsuperscript{*}Performance results of all baseline methods and the experimental setup are taken from \cite{singhal2025fed}. Trainable Parameter (TP) Efficiency Rank  quantifies the relative parameter efficiency of FedReFT(R8). Best results are in \textbf{bold}.}
\centering
\begin{tabular}{lcccccccccccc}
\toprule
\textbf{Method} & \textbf{R} & \textbf{TP(M) $\downarrow$} &
\textbf{BoolQ} & \textbf{PIQA} & \textbf{SIQA} &
\textbf{HellaS} & \textbf{WinoG} & \textbf{ARC-e} &
\textbf{ARC-c} & \textbf{OBQA} & \textbf{Avg$\uparrow$} \\
\midrule
FedIT\textsuperscript{*}        & 32  & 48.63  & 60.89 & 78.22 & 69.92 & 73.18 & 67.78 & 81.21 & 67.04 & 66.91 & 70.80 \\
FFA-LoRA\textsuperscript{*}     & 32  & 24.31  & 60.73 & 76.91 & 65.37 & 68.61 & 61.89 & 79.41 & 62.92 & 63.12 & 67.17 \\
FedEx-LoRA\textsuperscript{*}   & 32  & 243.15 & 62.55 & 79.36 & 71.41 & 71.78 & 72.45 & 82.69 & 67.80 & 70.25 & 73.13 \\
FLoRA\textsuperscript{*}        & 32  & 243.15 & 62.55 & 79.36 & 71.41 & 71.78 & 72.45 & 82.69 & 67.80 & 70.25 & 73.13 \\
Fed-SB\textsuperscript{*}  & 200 & 7.85  & 63.28 & 80.34 & 73.56 &
82.07 & 76.01 & 84.01 &
69.02 & 72.46 & 75.21 \\
\midrule
\multirow{2}{*} {\textbf{FedReFT(ours)}}
& 8 & 2.75 & 63.84 & 78.01 & 77.53 & 89.74 & 72.30 & 84.57 & 69.31 & 77.00 & 76.41 \\
 & \textbf{16} & \textbf{5.50} & \textbf{64.67} & \textbf{81.88} & \textbf{78.06} & \textbf{89.88} & \textbf{77.82} & \textbf{86.11} & \textbf{69.37} & \textbf{76.20} & \textbf{78.00} \\

\bottomrule
\end{tabular}

\label{tab:common_DT_nov_2025}
\end{table*}

\begin{table*}[htbp] 
\setlength{\tabcolsep}{4.6pt}
\centering
\caption{Performance comparison across arithmetic reasoning tasks with the Distinct Task (DT) and Mixed Task (MT) setup using different model sizes. We report results under adaptive mixing with Test-Time Computing, which dynamically balances local and global interventions at inference. Avg represents the average of the clients' accuracy.}
\begin{tabular}{c|cccc|cccc}
\toprule
\multicolumn{1}{c|}{\textbf{FedReFT}} & \multicolumn{4}{c|}{\textbf{Distinct Task (DT)}} & \multicolumn{4}{c}{\textbf{Mixed Task (MT)}} \\
\cmidrule(r){2-5} \cmidrule(l){6-9}
\textbf{Models}& \textbf{AQuA} & \textbf{SVAMP} & \textbf{MAWPS} & \textbf{Avg Acc$\uparrow$} & \textbf{AQuA} & \textbf{SVAMP} & \textbf{MAWPS} & \textbf{Avg Acc$\uparrow$} \\
\midrule
LLaMA 7B      &  26.12 & 26.71 & 49.94 & 34.26  & 20.86 & 15.60 & 28.22 & 22.23   \\
LLaMA-2 7B    &  30.96 & 33.31 & 59.51 & 41.93 & 22.05 & 23.50 & 32.74 & 26.09  \\
LLaMA-3 8B    &  35.36 & 49.41 & 75.25 & 53.34 & 33.45 & 51.28 & 73.51 & 52.75 \\

\bottomrule
\end{tabular}
\label{tbl:math_setup_1_2_resutls}
\end{table*}

\section{Experimental Validation}
\label{sec:experiments}



To evaluate FedReFT, we conduct extensive experiments on three different NLP benchmarks covering over 12 datasets. Our objective is to present a comprehensive assessment of how this approach performs in various NLP tasks. We experiment with both masked and autoregressive language models, including RoBERTa-large, TinyLLaMA-1B, LLaMA 7B, LLaMA-2 7B and 13B, LLaMA-3.2B and LLaMA-3 8B, across multiple settings and scales. Our comparisons include state-of-the-art baselines, such as LoRA\cite{hu2021lora}, FedIT \cite{zhang2024towards}, FFA-LoRA\cite{sun2024improving}, FedDPA-LoRA\cite{long2024dual}, FedSA-LoRA\cite{guo2024selective}, Fed-SB \cite{singhal2025fed} and FLoRA\cite{wang2024flora} focusing on both parameter efficiency and performance trade-offs. We align the experimental setup configurations with the baseline papers to ensure fair comparisons. To optimize memory usage, we load all base language models with \texttt{torch.bfloat16} precision. The results are averaged over two runs to report the mean performance. \\
\textbf{Hyperparameter Configuration.} In the experiments, we determine the number of interventions to learn and the specific layers and input positions where they are applied. Interventions are inserted at a fixed number of layers ($L$) and at the prefix ($p$) and suffix ($s$) positions of the input prompt. We narrow the hyperparameter search space for FL in Appendix \ref{appendix:hyperparameter} by adopting the configuration used in the centralized ReFT\cite{Wu2024} paper. \\
\textbf{Task Distribution Rationale.}
We design two experimental setups for commonsense and arithmetic reasoning tasks to study how global representations converge under diverse task distributions. In the Mixed-Task (MT) setup, each client trains on a subset of a combined reasoning dataset but is evaluated on a single task, encouraging generalized, transferable representations through ABM aggregation. This reflects collaborative learning across varied yet related tasks. In the Distinct Task (DT) setup, each client trains on a unique reasoning task, enabling personalized fine-tuning while still leveraging global updates. Despite higher task heterogeneity, this setup maintains stable performance as model capacity increases. Both setups show that FedReFT supports effective generalization in MT and robustness in DT.
\subsection{Commonsense Reasoning}
We evaluate global representation generation on eight commonsense reasoning tasks using the Commonsense170K dataset inspired by~\cite{singhal2025fed, Wu2024}. We use the same hyperparameter of ~\cite{singhal2025fed} and tune the intervention parameter in the Appendix \ref{appendix:hyperparameter_arithmetic_reasoning}.  This helps us tune important hyperparameters efficiently and also test their robustness across multiple commonsense reasoning tasks.\\
\textbf{Datasets.}
\label{ref:dataset_setup_experiments}
For the first setup, as a Mixed-Task (MT) design, we split the commonsense reasoning tasks dataset Commonsense170K \cite{hu2023llm} among clients and use them for fine-tuning. Each client evaluates one of the commonsense reasoning tasks. BoolQ, PIQA, SIQA, HellaSwag, and WinoGrande. For the second setup as Distinct Task (DT) design, each client fine-tunes on only one of these five commonsense reasoning tasks and is evaluated using the same task. We adopt the prompt template from Hu et al.~\cite{hu2023llm}. The Distinct Task (DT) experiment is described in the Appendix \ref{tbl:able_commonsense_DT}, as no baseline was found in this setting.
\begin{table*}[htbp]
\setlength{\tabcolsep}{2.2pt}
\renewcommand{\arraystretch}{1.0}
\caption{Performance comparison across GLUE Tasks on RoBERTa model for $C=3$, FedReFT use rank 1. \textsuperscript{*}Performance results of all baseline methods are taken from \cite{guo2024selective} and use LoRA rank 8. FedReFT achieves higher accuracy over all the baselines in FL settings. \textbf{(TP) Efficiency} quantifies the relative trainable parameter efficiency of FedReFT, indicating how many times fewer parameters it requires compared to baselines.}
\label{tbl:glue_roberta_large}
\begin{tabular}{llcccccccc}
\toprule
\textbf{Setup} & \textbf{Method} & \textbf{TP(M)$\downarrow$} & \textbf{TP Effi.  $\downarrow$}  & \textbf{MNLI-m} & \textbf{MNLI-mm} & \textbf{SST-2} & \textbf{QNLI} & \textbf{QQP} & \textbf{Avg $\uparrow$} \\
\midrule
\multirow{3}{*}{Standalone}
& Full Tuning  & 355 & 6698.11$\times$ & 88.8 & 88.56 & 96.0 &  93.8& 91.5&  91.73 \\
& LoRA\textsuperscript{*}     & 1.83 & 34.53$\times$ & 88.71 & 88.21 & 95.16 & 91.16 & 85.33 & 89.71 \\
& LoReFT    & 0.053 & 1.0$\times$ &  89.2& 89.26&  96.20 & 94.10 & 88.5& 91.45  \\
\midrule
\multirow{4}{*}{FL}
& FFA-LoRA\textsuperscript{*}  & 1.44 & 27.17$\times$ & 88.83  & 88.27 & 94.95 & 91.52 & 86.71 & 89.39 \\
& FedDPA-LoRA\textsuperscript{*} & 2.62 & 49.44$\times$ & 88.99  & 88.43 & 95.50 & 90.74 & 85.73 & 89.47 \\
& FedSA-LoRA\textsuperscript{*} & 1.83  & 10.40$\times$ & 90.18 & 88.88 & 96.00 & 92.13 & 87.48 & 90.43 \\
& \textbf{FedReFT} &  \textbf{0.053} & \textbf{1.0$\times$} & \textbf{89.75} & \textbf{89.31} & \textbf{95.75}  & \textbf{94.91}& \textbf{87.15}& \textbf{91.37}\\ 
\bottomrule
\label{tbl:glue_task}
\end{tabular}
\end{table*} \\
\textbf{Results.}
In Table~\ref {tbl:baseline_first_commmonsense}, our proposed FedReFT method demonstrates strong parameter efficiency while maintaining competitive accuracy across five commonsense reasoning tasks. Notably, FedReFT with rank 8 uses only 2.75M (0.0857\%) trainable parameters, achieving accuracy close to or better than several baselines. Compared to existing methods 
our approach reduces the trainable parameter count by factors of 1.03$\times$ to 13.68$\times$, with minimal to no compromise in performance, also shown in Figure \ref{fig:trainable_param}. The experiments results on the Mixed-Task (MT) setup are shown in Appendix Table \ref{tbl:second_setup_commonsense}.
\subsection{Arithmetic Reasoning}
For the arithmetic reasoning tasks, we design three experimental settings to fine-tune models on various arithmetic reasoning tasks. We follow the same hyperparameter tuning strategy as used in Commonsense170K in Appendix \ref{appendix:hyperparameter_arithmetic_reasoning}, which uses a development set to select the best-performing configuration. Evaluation is based solely on the final numeric or multiple-choice answer, disregarding intermediate reasoning steps. \\
\textbf{Datasets.} In the first setting following Mixed-Task (MT), we split the arithmetic reasoning dataset, MATH10K~\cite{hu2023llm}, which includes four tasks with chain-of-thought solutions generated by a language model. Each client reports performance using test set one of three tasks: AQuA, SVAMP, and MAWPS. In the second setting following Distinct-Task (DT), each client is assigned one arithmetic reasoning task for both fine-tuning and evaluation. In the third setting, following \cite{guo2024selective, kuang2024federatedscope}, we split the dataset GSM8K into 3 clients under an IID distribution. \\
\textbf{Results.}
In Table~\ref{tbl:math_setup_3_baseline}, FedReFT achieves the highest accuracy among all methods while requiring substantially fewer trainable parameters, underscoring its efficiency and superior performance. In Table~\ref{tbl:math_setup_1_2_resutls}, the DT setup represents task heterogeneity, enabling clients to learn personalized, task-specific representations while benefiting from global aggregation. Consequently, the DT setup achieves higher performance. In contrast, the Mixed-Task (MT) setup trains clients on heterogeneous task mixtures to promote globally generalizable representations, but this blending can reduce performance on specific evaluation tasks due to representation misalignment and conflicting objectives. 
\begin{table}[!htbp] 
\setlength{\tabcolsep}{0.3pt}
\renewcommand{\arraystretch}{.99} 
\centering
\caption{Performance comparison on arithmetic reasoning task for GSM8K on LLaMA-3 8B with rank 8, where clients enable consistent evaluation of representation generalization.\textsuperscript{*}Performance results of all baselines are taken from \cite{guo2024selective}. \textbf{(TP) Effi} measures how many times fewer trainable parameters it uses compared to the baselines, while $\Delta$\textbf{Acc} indicates the accuracy gain achieved by FedReFT over baseline.}
\begin{tabular}{lcccc}
\toprule
\textbf{Method} & \textbf{TP(M)$\downarrow$} & \textbf{TP Effi.$\downarrow$} & \textbf{Acc} & $\Delta$\textbf{Acc}$\uparrow$\\
\midrule
LoReFT  &  4.19 & 1.0$\times$ & 48.33 & +0.55  \\ 
LoRA\textsuperscript{*} &  30.40 & 7.26$\times$ & 46.23  & +2.65\\
\midrule
FedSA-LoRA\textsuperscript{*} &  30.40 & 7.26$\times$ & 46.63 & +2.25 \\
FFA-LoRA\textsuperscript{*} & 15.2 & 3.63$\times$ & 46.32 & +2.56\\
 \textbf{FedReFT(ours)} &  \textbf{4.19} & \textbf{1.0$\times$} & \textbf{48.88} & \\
\bottomrule
\label{tbl:math_setup_3_baseline}
\end{tabular}
\end{table}
\subsection{Natural Language Understanding}
We evaluate the effectiveness of FedReFT in learning generalizable representations for Natural Language Understanding (NLU) using the GLUE benchmark~\cite{wang2018glue}. The objective is to fine-tune NLU to learn global representations that capture task-level semantics. By aligning intermediate representations for downstream classification performance. This setup allows us to test whether lightweight intervention tuning can align representations across clients within a single NLU task.\\
\textbf{Results.} Table \ref{tbl:glue_task} depicts that our approach performs strongly across GLUE tasks while using very few trainable parameters. It performs competitively or surpasses other methods, demonstrating its ability to learn strong representations even under federated conditions. Despite utilizing 10$\times$–49$\times$ fewer trainable parameters than several baselines, it achieves comparable accuracy, underscoring its efficiency and effectiveness.
\subsection{Ablation Studies}
We conduct ablation studies in Appendix \ref{appendix:Ablation_study} to further evaluate the effectiveness of FedReFT, focusing on the impact of geometric-median-based All-But-Me aggregation, intervention parameter sharing strategies, and local model update approaches using balanced and TTC-based adaptive mixing
\section{Conclusion}
\label{sec:conclusion}
In this work, we bridge the gap between Representation Fine-Tuning (ReFT) and Federated Learning by introducing FedReFT, a unified framework that enables personalized and parameter-efficient federated representation learning. FedReFT employs the \textit{All-But-Me} aggregation strategy to mitigate semantic misalignment caused by naive averaging, enabling clients to adapt using a robust average of others’ interventions. Additionally, an adaptive mixing mechanism inspired by \textit{Test-Time Computing} dynamically balances local and global representations, enhancing robustness under heterogeneous conditions. Extensive experiments demonstrate that FedReFT consistently improves convergence, generalization, and parameter efficiency across diverse FL settings.
\section*{Limitations}
\label{appendix:limitations}
Due to computational constraints, our current study focuses primarily on LoReFT-based interventions within language models under a fixed set of hyperparameters. In future work, we aim to automate the parameter search space using a multi-agent coordination framework to better explore optimal low-rank configurations for each client. Although our current set-up does not explicitly address privacy, we are actively investigating how to integrate differential privacy mechanisms, such as DP-SGD, into the FedReFT framework without sacrificing personalization. Initial experiments in this direction are ongoing. Additionally, we are exploring the theoretical properties of ABM aggregation under adversarial or noisy clients, and whether it can be extended to other modalities beyond language, such as vision-language models in federated systems.
\label{ethical}
\subsection*{Data and Model Usage}
We use publicly available models including LLaMA-1.1B, LLaMA-2 (7B, 13B), LLaMA-3 8B, LLaMA-3.2 3B and RoBERTa-large. LLaMA-2 and LLaMA-3 models are licensed under Meta’s community license permitting commercial use. RoBERTa-large is under the MIT License, and TinyLLaMA use Apache 2.0, while the original LLaMA-1 7B is for non-commercial research only. We will release code and configurations under an open-source license with usage documentation to support reproducibility and responsible use. 

We employ publicly available datasets across commonsense and arithmetic reasoning tasks, each released under open-source licenses. For commonsense reasoning, BoolQ is under CC BY-SA 3.0, PIQA under Apache 2.0, SIQA and WinoGrande under CC BY 4.0, HellaSwag under MIT, ARC under CC BY-SA 4.0, and OBQA under CC BY 4.0. For arithmetic reasoning, AddSub, AQuA, MAWPS, and MultiArith are under Apache 2.0, GSM8K and SVAMP under MIT, and SingleEq under CC BY 4.0. For natural language understanding, GLUE consists of multiple datasets, each with its own license, allowing for research use and redistribution. 

\subsection*{Environmental Impact}
Our approach  FedReFT achieves 1$\times$– 49$\times$ higher parameter efficiency than existing PEFT methods, using fewer trainable parameters. This reduces energy consumption and training time, making our method more resource-efficient and environmentally friendly.

\subsection*{Societal Impacts}
Our method FedReFT adapts ReFT for Federated Learning, enabling efficient model personalization with minimal computational overhead. This promotes broader accessibility of large language models on edge devices, including in low-resource or privacy-sensitive environments. While improving inclusivity and deployment scalability, care must be taken to mitigate potential misuse or bias propagation across decentralized systems.

\subsection*{Bias and Fairness}
Our approach FedReFT considers the potential for bias introduced by non-IID client data in Federated Learning. While we do not explicitly optimize for fairness, we acknowledge that imbalanced participation or data diversity may lead to uneven model performance. Future work should explore fairness-aware objectives to mitigate such disparities across clients and demographic groups.

\subsection*{Responsible Deployment}
To support responsible use, we include clear documentation outlining the intended use cases of our framework and advise against applying it in safety-critical settings without thorough validation. We encourage users to follow ethical standards, such as the ACL Code of Ethics, when deploying our method. Our released code comes with usage instructions to promote safe adoption and reduce the risk of misuse. This work is licensed under CC BY 4.0, allowing reuse and adaptation, even commercially, with proper attribution.

\subsection*{AI Assistants in Research Writing}
We used AI assistants to support writing and code refinement during the preparation of this paper. All AI-generated content was reviewed and verified by the authors.

\bibliography{main}

\appendix
\section{Related Works}
\label{appendix:related}
\subsection{Parameter-Efficient Fine-Tuning (PEFT)}
Fine-tuning LLMs is resource-intensive due to their large parameter counts. Parameter-efficient fine-tuning (PEFT) methods mitigate this by updating only a small subset of parameters while keeping pre-trained weights frozen~\cite{li2021prefix,he2021towards,wang2022adamix}. Several PEFT approaches have been proposed, Adapter Tuning~\cite{houlsby2019parameter}, BitFit~\cite{zaken2022bitfit}, Prefix Tuning~\cite{li2021prefix}, Prompt Tuning~\cite{lester2021power}, and Low-Rank Adaptation (LoRA)~\cite{Hu21}. Among them, LoRA is widely adopted for its efficiency in approximating weight updates via low-rank matrices. Extensions such as ReLoRA~\cite{lialin2023relora} and RankAdapter~\cite{zhou2024rankadaptor} improve memory use and adapt ranks dynamically, though they lack theoretical guarantees. AdaZeta~\cite{yang2024adazeta} introduces zeroth-order optimization with convergence guarantees, while others~\cite{gao2024adaptive, rajabzadeh2024qdylora, valipour2022dylora} explore adaptive ranks without formal proofs. LoRA has been integrated with Mixture-of-Experts models~\cite{li2024mixlora, wu2024mixture}, as in AdaMoLE~\cite{liu2024adamole}, to enable dynamic expert selection, and also with Neural Architecture Search for LLM compression \cite{munoz2025low}. These approaches primarily target weight updates, overlooking direct interventions in hidden representations, which are discussed next.

\subsection{Representation Fine-Tuning (ReFT)}
ReFT shifts fine-tuning from model weights to hidden representations, leveraging their semantic structure for efficient adaptation~\cite{Wu2024}. Inspired by activation steering and representation engineering~\cite{avitan2024changed, li2024inference, liu2023context, singh2024mimic}, ReFT enables task-specific control through fixed or learned interventions without updating the full model. Notably, Inference-Time Intervention (ITI)~\cite{li2024inference} improves LLM truthfulness by modifying activations, while representation engineering~\cite{zou2023representation} combines representation reading and control for interpretable model behavior. Minimally Modified Counterfactuals (MMC)~\cite{singh2024mimic} unify erasure and steering to reduce bias, and can be mapped to natural language edits~\cite{avitan2024changed}, enhancing interpretability. These findings support direct representation manipulation as a lightweight and effective alternative to weight-based PEFT methods like LoRA.

\subsection{Federated Fine-Tuning}
Federated Learning (FL)~\cite{mcmahan2017communication} poses challenges for fine-tuning LLMs, including data heterogeneity, communication constraints, and model diversity. PEFT methods have emerged to address these issues efficiently~\cite{sun2022exploring, chen2022fedtune, zhang2023fedpetuning}. LoRA-based approaches such as FedLoRA~\cite{yi2023fedlora}, Hyper-FloRA~\cite{lu2024hyperflora}, and Efficient FL Adapter~\cite{cai2023efficient} offer modular and personalized adaptation across clients. Recent advances further incorporate privacy (FFA-LoRA~\cite{sun2024improving}), heterogeneous adaptation (FloRA~\cite{wang2024flora}), instruction tuning (FedIT~\cite{zhang2024towards}), residual learning in FRLoRA ~\cite{yanfederated} which tackles client drift by directly adding residual low-rank weight products to the global model parameters in each round, and heterogeneous resources in FlexLoRA ~\cite{bai2024federated} which mitigates the bucket effect by leveraging SVD to aggregate and redistribute LoRA weights with varying ranks among clients, and model compression in FedBiOT~\cite{wu2024fedbiot} which enables LLM fine-tuning without requiring clients to access the full model by using a bi-level optimization scheme to align a compressed emulator with a lightweight adapter, and adaptive sketching in FSLoRA ~\cite{fang2025federated} which reduces communication and computation overhead by allowing clients to selectively update submatrices of the global LoRA modules based on sketching ratios, and expert routing (DualFed~\cite{long2024dual}, Sparse-FedMoE~\cite{tran2025revisiting}). In contrast, our proposed \textsc{FedReFT} shifts from weight updates to direct representation-level tuning via sparse intervention layers and introduces an All-But-Me (ABM) aggregation strategy to preserve semantic alignment while enabling robust knowledge sharing across non-IID clients.

\subsection{Aggregation Methods in FL}
To address the inherent heterogeneity and robustness challenges in federated learning, median-based aggregation strategies have been extensively studied as alternatives to simple averaging. Unlike the arithmetic mean, the geometric and coordinate-wise medians are significantly more resilient to outliers and adversarial updates, making them suitable for secure and personalized FL scenarios. For instance, coordinate-wise median aggregation has been proposed to defend against Byzantine clients in distributed optimization \cite{blanchard2017machine}. This was extended with geometric median-based gradient descent to improve statistical guarantees across diverse loss landscapes \cite{yin2018byzantine}. Further work demonstrated that coordinate-wise median and trimmed-mean-based methods achieve order-optimal convergence not only for strongly convex losses but also under non-strongly convex and even non-convex population losses \cite{chen2017distributed}. Additionally, a one-round median-based algorithm was shown to maintain statistical optimality under quadratic convexity, offering a communication-efficient solution \cite{chen2017distributed}. RFA \cite{pillutla2022robust} maintains privacy and demonstrates improved robustness over standard averaging techniques, particularly in environments with high levels of data corruption. FedEAT \cite{pang2025fedeat} integrates adversarial training in the embedding space with geometric median-based aggregation to enhance robustness while preserving performance. This work demonstrates that LoRA-based FL systems can effectively leverage geometric median aggregation.
Inspired by these findings, we adopt geometric median aggregation in our FL framework to aggregate the All-But-Me (ABM) intervention parameter, weight $\mathbf{W}$, rotation $\mathbf{R}$, and bias $\mathbf{b}$. This provides stability across diverse client behaviors and loss geometries, improving personalization performance under data and objective heterogeneity.

\section{Hyperparameter Search Space}
\label{appendix:hyperparameter}
\subsection{Hyperparameter Search Space for Commonsense and Arithmetic Reasoning}
\label{appendix:hyperparameter_arithmetic_reasoning}

Following the ReFT framework~\cite{Wu2024}, we construct a development set using the GSM8K dataset and consider only the last 300 samples. We trained the clients using LLaMA 7B model with the remaining training data and determined the best-performing hyperparameters based on the model's performance on the development set. We further use this hyperparameter in another model directly. We set the maximum input sequence length to 512 tokens during training and tuning, and limit inference to 32 generated tokens. We use the same setup for commonsense reasoning with Commonsense170K dataset. The hyperparameter search space is summarized in  Tables~\ref{tab:appendix_gsm8k_hypertune} and~\ref{tab:appendix_commonsense_hypertune}.

During inference, we use greedy decoding (without sampling) for the commonsense reasoning benchmark, as it is a multi-token classification task. For arithmetic reasoning, we follow the decoding setup from ~\cite{hu2023llm}, using a higher temperature of 0.3. This change helps avoid errors in HuggingFace's decoding caused by unstable probabilities
\begin{table}[!htbp]
\centering
\caption{Narrow down the hyperparameter(HP) search space of LLaMA 7B models with FedReFT on the GSM8K development set, inspired from \cite{Wu2024}. The best-performing settings are \underline{underlined}. We apply greedy decoding without sampling during hyperparameter tuning.}
\begin{tabular}{ll}
\toprule
\textbf{HP} & \textbf{FedReFT} \\
\midrule
prefix+suffix, $p + s$ & \{p5+s5, \underline{p7+s7}, p9+s9\} \\
Tied weight $\phi$& \{True, \underline{False}\} \\
Rank $r$ & \{\underline{8}, 16, 32, 64\} \\
Layer $L$ & \{\underline{all}\} \\
Dropout & \{\underline{0.00}, 0.05\} \\
Optimizer & AdamW \\
LR & 
\{6, \underline{9}\}$\times$10$^{-4}$
\\
Weight decay & \{\underline{0}, 1$\times$10$^{-3}$, 2$\times$10$^{-3}$\} \\

LR scheduler & Linear \\
Batch size & \{\underline{16}, 32\} \\
Warmup ratio & \{0.06, \underline{0.10}\} \\
Clients & \{3, 5\} \\
Epochs & \{3, 4, \underline{5}, 6\} \\
Rounds & {10}  \\
\bottomrule
\label{tab:appendix_gsm8k_hypertune}
\end{tabular}
\end{table}
\begin{table}[!htbp]
\centering
\caption{Narrow down the hyperparameter (HP) search space of LLaMA 7B models with FedReFT on the Commonsense170K development set, following the Appendix \ref{appendix:hyperparameter_arithmetic_reasoning}. The best-performing settings are \underline{underlined}. We apply greedy decoding without sampling during hyperparameter tuning.}
\begin{tabular}{ll}
\toprule
\textbf{HP} & \textbf{FedReFT} \\
\midrule
prefix+suffix,  $p + s$ & \{p5+s5, \underline{p7+s7}\} \\
Tied weight $p, s$ & \{True, \underline{False}\} \\
Rank $r$ & \{\underline{8}, 16, 32, 64\} \\
Layer $L$ & \{\underline{all}\} \\
Dropout & \{\underline{0.00}, 0.05\} \\
Optimizer & AdamW \\
LR & \{4, \underline{6}, 9\}$\times$10$^{-4}$
\\
Weight decay & \{\underline{0}\} \\
LR scheduler & Linear \\
Batch size & \{\underline{16}, 32\} \\
Warmup ratio & \{\underline{0.1}\} \\
Clients & \{3, 5\} \\
Epochs & \{2, \underline{3}, 4\} \\
Rounds & {10}  \\
\bottomrule
\label{tab:appendix_commonsense_hypertune}
\end{tabular}
\end{table}
\subsection{Hyperparameter Search Space for GLUE Benchmark}
\label{appendix:hyperparameter_glue}
We perform hyperparameter tuning following common practice for PEFT methods \cite{hu2023llm} in FL on RoBERTa separately for each GLUE task in Table \ref{tab:appendix_roberta_hyper}, selecting the optimal settings based on validation performance using a fixed random seed of 42. Final evaluations are conducted using two additional unseen seeds, \{43, 44\}, to ensure robustness.
\begin{table}[!htbp]
\setlength{\tabcolsep}{4pt}
\centering
\caption{Hyperparameter(HP) settings of RoBERTa-large models on selected GLUE tasks for FedReFT, inspired from \cite{Wu2024}}
\begin{tabular}{lllll}
\toprule
\textbf{HP} & \textbf{MNLI} & \textbf{SST-2} & \textbf{QNLI} & \textbf{QQP} \\
\midrule
position $p$ & $p1$ & $p3$ & $p11$ & $p11$ \\
Tied weight & \multicolumn{4}{c}{False} \\
Rank $r$ & \multicolumn{4}{c}{1} \\
Layer $L$ & \multicolumn{4}{c}{all} \\
Dropout & \multicolumn{4}{c}{0.05} \\
Optimizer & \multicolumn{4}{c}{AdamW} \\
LR & \multicolumn{4}{c}{ $6\times 10^{-4}$} \\ 
Weight decay & \multicolumn{4}{c}{0.00} \\
LR scheduler & \multicolumn{4}{c}{Linear} \\
Batch size & \multicolumn{4}{c}{32} \\
Warmup ratio & 0.00 & 0.10 & 0.10 & 0.06 \\
Epochs & \multicolumn{4}{c}{5} \\ 
Rounds & \multicolumn{4}{c}{50} \\
\bottomrule
\label{tab:appendix_roberta_hyper}
\end{tabular}
\end{table}
\section{Theoretical Foundation: Geometric Median via Weiszfeld’s Algorithm}
\label{appendix:proofs}
The geometric median offers a robust alternative to the arithmetic mean, particularly suitable for federated settings with heterogeneous or noisy client updates. For a given set of vectors \(\mathcal{X} = \{\mathbf{x}_1, \mathbf{x}_2, \dots, \mathbf{x}_n\} \subset \mathbb{R}^d\), the geometric median \(\mathbf{y}^*\) is defined as:
\begin{equation}
\mathbf{y}^* = \arg\min_{\mathbf{y} \in \mathbb{R}^d} \sum_{i=1}^n \|\mathbf{y} - \mathbf{x}_i\|_2.
\end{equation}
This optimization is non-smooth and convex, and generally lacks a closed-form solution. However, Weiszfeld’s algorithm \cite{weiszfeld1937point} provides an efficient iterative method to approximate \(\mathbf{y}^*\). We now derive and justify this algorithm via the Majorization-Minimization (MM) framework.

We define the cost function to be minimized:
This function is convex but non-differentiable at points where \(\mathbf{y} = \mathbf{x}_i\). Weiszfeld’s algorithm avoids such points during updates by construction.

The MM algorithm minimizes a difficult objective \(f(\mathbf{y})\) by iteratively minimizing a surrogate function \(Q(\mathbf{y}|\mathbf{y}^{(k)})\) that:
Majorizes \(f\): \(Q(\mathbf{y}|\mathbf{y}^{(k)}) \geq f(\mathbf{y})\) for all \(\mathbf{y}\),
Touches \(f\) at the current iterate: \(Q(\mathbf{y}^{(k)}|\mathbf{y}^{(k)}) = f(\mathbf{y}^{(k)})\).

We define the surrogate using Jensen’s inequality and the convexity of the norm:

\begin{equation}
Q(\mathbf{y}|\mathbf{y}^{(k)}) = \sum_{i=1}^n \frac{\|\mathbf{y} - \mathbf{x}_i\|_2^2}{2 \|\mathbf{y}^{(k)} - \mathbf{x}_i\|_2} + C(\mathbf{y}^{(k)}),
\end{equation}

where \(C(\mathbf{y}^{(k)})\) is a constant that does not depend on \(\mathbf{y}\). This function is differentiable and strictly convex in \(\mathbf{y}\).

To find the minimizer of \(Q(\mathbf{y}|\mathbf{y}^{(k)})\), we take the gradient and set it to zero:
\begin{equation}
\nabla Q(\mathbf{y}) = \sum_{i=1}^n \frac{\mathbf{y} - \mathbf{x}_i}{\|\mathbf{y}^{(k)} - \mathbf{x}_i\|_2} = 0.
\end{equation}
Solving the above yields the Weiszfeld update rule:
\begin{equation}
\mathbf{y}^{(k+1)} = \frac{\sum_{i=1}^n \frac{\mathbf{x}_i}{\|\mathbf{y}^{(k)} - \mathbf{x}_i\|_2}}{\sum_{i=1}^n \frac{1}{\|\mathbf{y}^{(k)} - \mathbf{x}_i\|_2}}.
\end{equation}
The update is only valid when \(\mathbf{y}^{(k)} \neq \mathbf{x}_i\) for all \(i\), a condition that can be enforced by initialization and step-size dampening if needed.
From MM theory~\cite{lange2000optimization}, each iteration satisfies:
\begin{equation}
\begin{aligned}
f(\mathbf{y}^{(k+1)}) &\leq Q(\mathbf{y}^{(k+1)}|\mathbf{y}^{(k)}) \\
&\leq Q(\mathbf{y}^{(k)}|\mathbf{y}^{(k)}) = f(\mathbf{y}^{(k)}),
\end{aligned}
\end{equation}
ensuring that \(f(\mathbf{y}^{(k)})\) is non-increasing. Under mild conditions (excluding cases where \(\mathbf{y}^{(k)} = \mathbf{x}_i\)), Weiszfeld’s algorithm converges to the geometric median \(\mathbf{y}^*\).
\subsection{All-But-Me (ABM) Aggregation Strategy}
In our FedReFT framework, each client receives an All-But-Me (ABM)aggregated update for intervention parameters computed as the geometric median of the corresponding parameters from all other clients. For client \(k\), the ABM aggregated parameter is:
\begin{equation}
\mathbf{W}^{\text{ABM}}_k = \arg\min_{\mathbf{w} \in \mathbb{R}^d} \sum_{m \ne k} \|\mathbf{w} - \mathbf{W}_m^{\text{local}}\|_2.
\end{equation}
We compute this using Weiszfeld’s algorithm for each parameter type independently, ensuring robustness to outlier clients and misaligned updates. This enables stable and personalized aggregation without sacrificing task-specific semantics.
Weiszfeld’s algorithm provides a theoretically grounded and computationally efficient way to compute the geometric median, making it ideal for ABM aggregation in heterogeneous FL. By leveraging this algorithm in FedReFT, we ensure robustness in aggregation and improve both convergence and personalization in non-i.i.d. federated environments.

\subsection{Geometric Median over mean}
The Geometric Median used in All-but-Me (ABM) aggregation is computationally more expensive, with a time complexity of $\mathbf{O(T \cdot d)}$ due to its iterative Weiszfeld optimization and the need to accumulate all $N$ client updates. In this paper, the dimension is only $d=20$, which makes the overhead relatively small. In contrast, the arithmetic mean used in FedAvg has a lower complexity of $\mathbf{O(d)}$. However, the slightly higher cost of  the Geometric Median is justified by its stronger robustness: by minimizing the $\mathbf{L_1}$ error, it becomes significantly more resilient to client heterogeneity  and outlier updates. This leads to consistently higher accuracy, with clear performance  improvements observed over FedAvg on the Commonsense task. While outlier clien updates can render the aggregated FedAvg model unstable or even unusable, the Geometric Median ensures  a stable and reliable update, producing a more accurate final model.

\begin{figure}[htbp]
    \centering
    \includegraphics[width=1\linewidth]{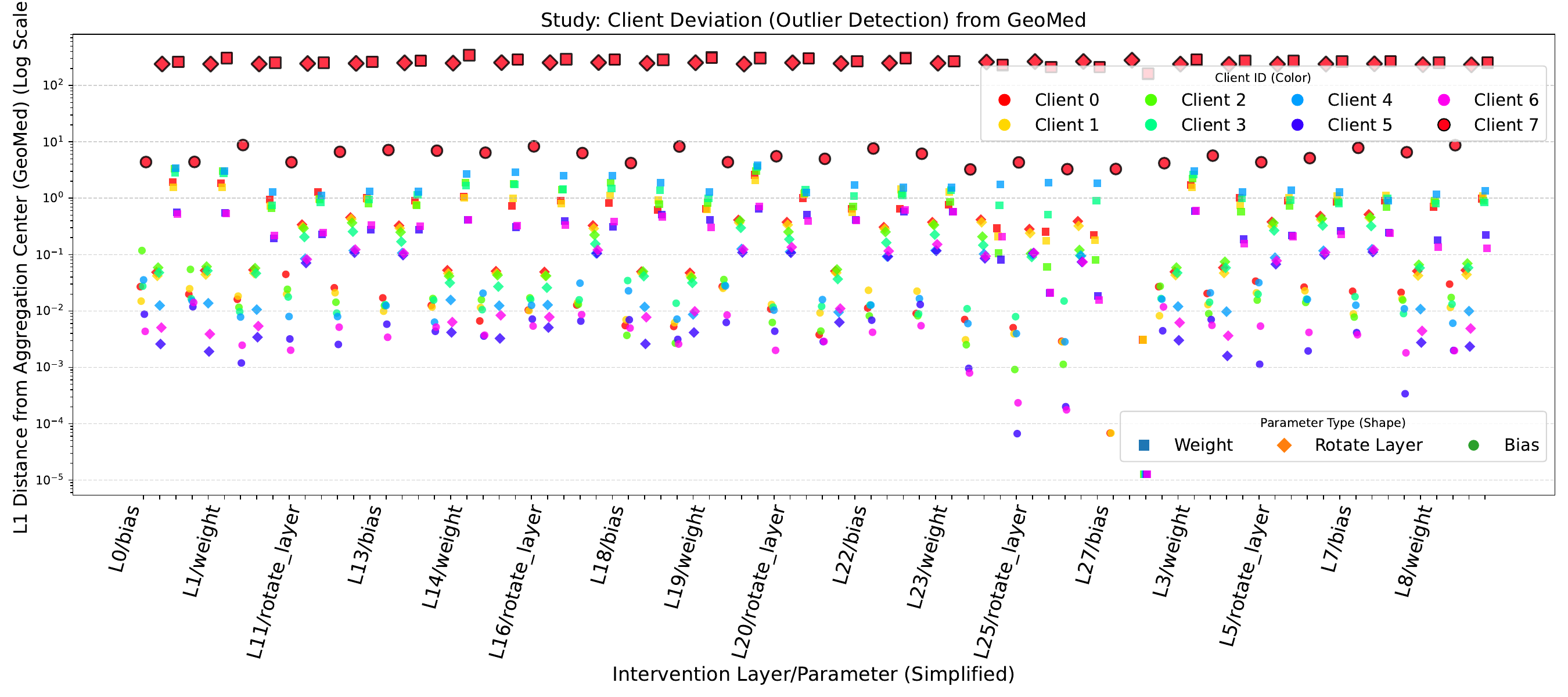}
    \caption{Study\_Distance\_from\_GeoMed\_ShapeColor}
    \label{fig:Study_Distance_from_GeoMed_ShapeColor}
\end{figure}

\begin{figure}[htbp]
    \centering
    \includegraphics[width=1\linewidth]{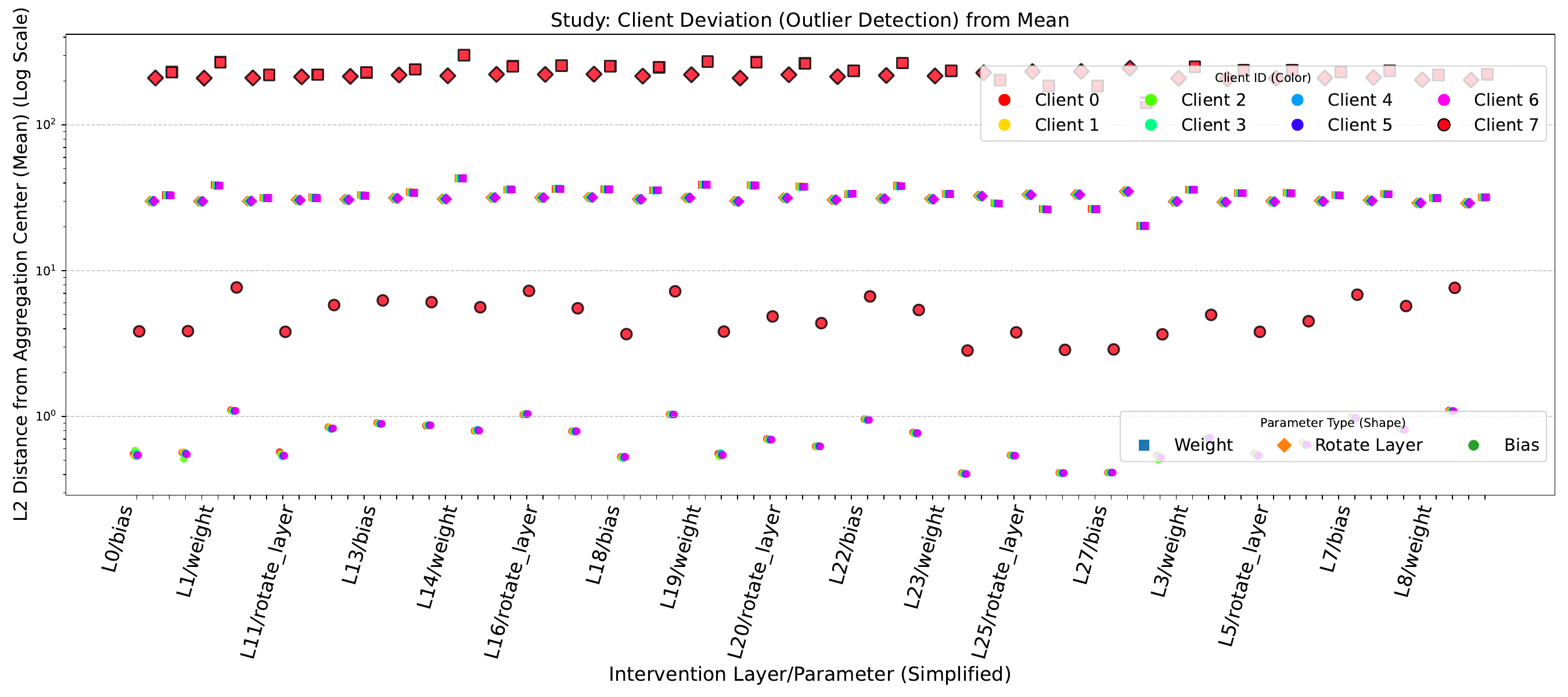}
    \caption{Study\_Distance\_from\_Mean\_ShapeColor}
    \label{fig:Study_Distance_from_Mean_ShapeColor}
\end{figure}

For a heterogeneous task setting of a commonsense task using 8 clients, the Geometric Median provides superior outlier detection in Figure \ref{fig:Study_Distance_from_GeoMed_ShapeColor}. The Distance from GeoMed plot visually proves Client 7 is an extreme outlier, clearly separating it from the benign majority's distance. This maximum separation occurs because GeoMed anchors near the majority by $\mathbf{L}_1$ linear error minimization, resisting the outlier's pull. Conversely, the Mean plot shows the Mean center is contaminated, minimizing the outlier's distance because the Mean minimizes $\mathbf{L}_2$ squared error in figure \ref{fig:Study_Distance_from_Mean_ShapeColor}. This distortion makes the aggregated model unusable, but GeoMed's robust calculation ensures a stable, accurate final update.

\section{Ablation Study}
\label{appendix:Ablation_study}
\subsection{Intervention Parameter Sharing Strategy}
\label{abl:Intervention_Parameter_Sharing_Strategy}
To reduce communication overhead while maintaining personalization, we explore three strategies for sharing local intervention parameters with the server. These strategies represent different trade-offs between expressiveness and communication efficiency:
\begin{itemize}
    \item \textbf{Full Intervention Sharing}: \{$\mathit{W} \in \mathbb{R}^{r \times d}$, $\mathit{R} \in \mathbb{R}^{r \times d}$, $\mathit{b} \in \mathbb{R}^{r}$\}
This strategy shares the complete set of intervention parameters, capturing client-specific compression ($\mathit{W}$), transformation ($\mathit{R}$), and translation ($\mathit{b}$). It enables the most accurate reconstruction of local updates and yields the best global performance, especially under high heterogeneity.
\item \textbf{No Bias Sharing}: \{$\mathit{W} \in \mathbb{R}^{r \times d}$, $\mathit{R} \in \mathbb{R}^{r \times d}$\}
This variant omits the bias term $\mathit{b}$ but retains the directional transformation via $\mathit{W}$ and $\mathit{R}$. While it allows the server to align low-rank subspace transformations across clients, it lacks the ability to model per-dimension translation shifts, which can hinder fine-grained personalization. 
\item  \textbf{No $\mathit{W}$ Sharing}: \{$\mathit{R} \in \mathbb{R}^{r \times d}$, $\mathit{b} \in \mathbb{R}^{r}$\}. This configuration excludes $\mathit{W}$, giving the server access only to the reconstruction and shift parameters. Without knowledge of how the local signals were encoded, the server's ability to interpret or align updates is severely limited. 
\end{itemize}
The $\{\mathit{W}, \mathit{R}, \mathit{b}\}$ strategy provides the highest fidelity for aggregation, $\{\mathit{W}, \mathit{R}\}$ offers a balanced compromise, and $\{\mathit{R}, \mathit{b}\}$ prioritizes communication efficiency at the cost of semantic alignment and global performance.
\begin{table}[!htbp]
\setlength{\tabcolsep}{4.5pt}
\centering
\caption{Performance vs. parameter efficiency for different LoReFT sharing strategies (Uplink) for $C$ clients on commonsense reasoning task following the second experiment design.}
\begin{tabular}{llcc}
\toprule
\textbf{Task} & \textbf{Strategy} & \textbf{TP(\% $\downarrow$)} & \textbf{Accu$\uparrow$} \\
\midrule
GLUE 
&\texttt{W,R,b} & 0.01384 & 94.91  \\
ROBERTa&\texttt{W,R}    & 0.01383 & 64.03  \\
&\texttt{R,b}    & 0.00693 &  74.12  \\
\midrule
Arithmetic 
&\texttt{W,R,b} & 0.03114 &  33.31   \\
LLaMA-2 7B&\texttt{W,R}    & 0.03114 & 26.13  \\
&\texttt{R,b}    & 0.01557 & 25.77 \\
\midrule
Commonsense
&\texttt{W,R,b} & 0.03114 & 76.55 \\
LLaMA-2 7B&\texttt{W,R}   & 0.03114 & 70.63  \\
&\texttt{R,b}   & 0.01557 & 68.02 \\
\bottomrule
\end{tabular}
\label{tab:ablation_uplink_strategy}
\end{table}

\subsection{Partial Client Participation (PCP)}
The ability of a federated learning system to maintain performance when only a fraction of clients participate in each round, known as Partial Client Participation (PCP), is crucial for real-world deployment. This robustness to PCP demonstrates the system's ability to handle practical constraints like communication bandwidth limitations, client availability, and device battery life. To evaluate partial client participation (PCP), the experiment's results in Table \ref{tbl:glue_task_pcp} use a total of $C = 5$ clients. When all five clients participated in every global round, FedReFT obtained an average accuracy of $91.45\%$. Under PCP, where only $C = 3$ clients were randomly selected per round, FedReFT achieved an average accuracy of $90.62\%$. The small difference between the two settings indicates that FedReFT maintains stable performance even with reduced client participation. 
\begin{table}[!h]
\setlength{\tabcolsep}{3pt}
\caption{Partial client participation (PCP) and all clients in FL settings for GLUE Task for FedReFT}
\label{tbl:glue_roberta_large_pcp}
\begin{tabular}{llccccccc}
\toprule
\textbf{Method}  & \textbf{MNLI-m} & \textbf{SST-2} & \textbf{QNLI} & \textbf{QQP} & \textbf{Avg $\uparrow$} \\
\midrule
PCP & 89.17 & 92.83& 93.81 & 86.65 & 90.62\\
All & 89.61  & 95.25 & 94.02 & 86.93 & 91.45 \\
 \bottomrule
\label{tbl:glue_task_pcp}
\end{tabular}
\end{table} 

\subsection{Integrating Test-Time Computing (TTC) with Adaptive Mixing in Local Model Updates}
\label{abl:TTC_VS_Balanced_mixing}

In FedReFT, local interventions encode client-specific semantics, while ABM aggregation conveys shared global knowledge. However, naively averaging representation-level updates can disrupt semantic consistency across clients, leading to misaligned feature spaces. Furthermore, uniform aggregation disregards client heterogeneity, reducing adaptability in diverse edge environments. 

To address these challenges, we evaluate the effectiveness of FedReFT under two mixing strategies: (i) adaptive mixing via Test-Time Computing (TTC), where local and global interventions are dynamically balanced, and (ii) balanced mixing, where a fixed coefficient $\alpha$ governs the trade-off between local ($\alpha$) and global ($1-\alpha$) contributions. While TTC provides a dynamic, task-adaptive mechanism expected to outperform balanced mixing ($\alpha = 0.5$), we also include results for the fixed $\alpha = 0.5$ setting to illustrate this trade-off. Table \ref{tbl:baseline_first_commmonsense_ablation}, \ref{tbl:glue_task_ablation}, \ref{tbl:math_setup_1_2_resutls_ablation}, and \ref{tbl:math_setup_3_baseline_ablation} show that the performance of TTC-based adaptive mixing is over the balanced mixing. 

In equation \ref{eq:arg_min_lamda}, the Test-Time Computing (TTC) optimization uses three small regularization weights, $\lambda_1, \lambda_2, \lambda_3$, all set to 0.001. This low value means the primary goal is minimizing the test loss $\mathcal{L}_{\text{test}}$, not strict following of the regularization rules. $\lambda_1$ Entropy and $\lambda_3$ Diversity gently stop the mixing coefficient $\alpha$ from completely favoring local (1) or global (0) updates. $\lambda_2$ Consistency mildly allows $\alpha$ to change across different intervention layers, which helps with adaptive personalization. Because the $\lambda$ values are small, the resulting mean $\alpha \approx 0.57$ strongly favors local updates based on the client's specific data, but not the fully Local-Only assumption of $\mathbf{\alpha=}$1.0. If these $\lambda$ values were increased significantly, the regularization would become more important than the test loss. This would push $\alpha$ closer to $0.5$ for balanced mixing across all layers, reducing the model's ability to personalize. 
\begin{table*}[!htbp]
\setlength{\tabcolsep}{2.1pt}
\centering
\caption{Federated fine-tuning performance of LLaMA-3.2 3B across five commonsense reasoning tasks with the Mixed-Task (MT) setup, where clients train on heterogeneous task mixtures. 
We report the effectiveness of \textbf{FedReFT} under two mixing strategies: (i) \textbf{adaptive mixing} using Test-Time Computing (TTC), where local and global interventions are dynamically balanced, and (ii) \textbf{balanced mixing}, where a coefficient $\alpha$ controls the trade-off ($\alpha$ for local and $1-\alpha$ for global). 
While TTC is intended to provide a dynamic, task-adaptive solution expected to surpass balanced mixing ($\alpha=0.5$), we also include results for $\alpha = 0.5$ to illustrate the trade-off. 
\textbf{R} = Rank, \textbf{TP(M)} = Trainable Parameters in Millions, and \textbf{TP Effi. (R8)} = parameter efficiency of FedReFT with rank 8 compared to baselines. 
Baseline numbers are taken from \cite{singhal2025fed}.}
\label{tab:fed-sb-performance}
\begin{tabular}{lccccccccc}
\toprule
\textbf{Method} & \textbf{R} & \textbf{TP(M) $\downarrow$} & \textbf{TP Effi. (R8)} $\downarrow$ & \textbf{BoolQ} & \textbf{PIQA} & \textbf{SIQA} & \textbf{HellaS.} & \textbf{WinoG} & \textbf{Avg Acc$\uparrow$}\\
\midrule
FedIT\textsuperscript{*}         & 32  & 48.63  & 17.68$\times$ & 62.99 & 81.50 & 73.13 & 76.83 & 71.51 & 75.74 \\
FFA-LoRA\textsuperscript{*}      & 32  & 24.31  & 8.84$\times$ & 62.87 & 80.03 & 68.53 & 70.02 & 65.56 & 71.11 \\
Fed-SB \textsuperscript{*}       & 120 & 2.83 & 1.03$\times$ &64.86 & 81.66 & 74.87 & 81.67 & 75.22 & 75.66 \\ 

\midrule
\multirow{2}{*} {\textbf{FedReFT (TTC)}}
 & 4 & 1.38 & 0.5$\times$ & 63.35 & 82.72 & 72.96 & 91.37 & 69.70 & 76.02 \\
 & \textbf{8} & \textbf{2.75} &  \textbf{1.0$\times$} & \textbf{65.50} & \textbf{82.32} & \textbf{73.28} & \textbf{91.43} & \textbf{70.24} & \textbf{76.55}\\

\midrule
\multirow{4}{*} {\textbf{FedReFT ($\alpha$=0.5) }}
  & 4 & 1.38 &  0.5$\times$ & 63.09 & 82.10 & 72.36 & 90.27 & 69.22 & 75.41 \\
 & 8 & 2.75 & 1.0$\times$ & 64.01 & 81.18 & 72.11 & 90.71 & 71.01 & 75.66 \\ 
 & 16 & 5.5 &  $0.5\times$ & 63.42 & 81.61 & 73.64 & 91.23 & 71.35 & 76.05 \\
 & 32 & 11.01 &  $0.25\times$&  64.53  & 81.34 & 73.39 & 91.51 & 71.32 & 76.22 \\


 
 \midrule
\multirow{2}{*} {\textbf{FedReFT (tie $\phi$)}}
& 4 & 0.688 &  0.0214 & 49.94 & 81.23 & 72.72 & 89.84 & 68.43 & 72.43 \\
& 8 & 1.38 & 0.0428 & 57.15 & 81.22 & 72.77 &  90.56 &  68.50 & 74.04 \\ 
\bottomrule
\label{tbl:baseline_first_commmonsense_ablation}
\end{tabular}
\end{table*}

\begin{table*}[!htbp]
\setlength{\tabcolsep}{1.8pt}
\renewcommand{\arraystretch}{1}
\caption{Performance comparison across GLUE tasks on RoBERTa with $C=3$. 
We report results under two mixing strategies: (i) \textbf{adaptive mixing} with Test-Time Computing (TTC), which dynamically balances local and global interventions at inference, and (ii) \textbf{balanced mixing}, where $\alpha$ denotes the proportion of local contribution and $1-\alpha$ the global contribution. 
The balanced $\alpha = 0.5$ cases illustrate trade-offs between local personalization and global knowledge, while TTC is intended as a dynamic alternative expected to generalize better across heterogeneous conditions. \textbf{TP} denotes the number of trainable parameters (in millions), and \textbf{TP Effi.} measures the parameter efficiency of FedReFT (rank 1) relative to all LoRA-based baselines \cite{guo2024selective} with rank 8.}
\label{tbl:glue_roberta_large}

\begin{tabular}{llcccccccc}
\toprule
\textbf{Setup} & \textbf{Method} & \textbf{TP(M) $\downarrow$} & \textbf{TP Effi.$\downarrow$}  & \textbf{MNLI-m} & \textbf{MNLI-mm} & \textbf{SST-2} & \textbf{QNLI} & \textbf{QQP} & \textbf{Avg $\uparrow$} \\
\midrule
\multirow{3}{*}{Standalone}
& Full Tuning  & 355 & 6698.11$\times$ & 88.8 & 88.56 & 96.0 &  93.8& 91.5&  91.73 \\
& LoRA\textsuperscript{*}     & 1.83 & 34.53$\times$ & 88.71 & 88.21 & 95.16 & 91.16 & 85.33 & 89.71 \\

& LoReFT    & 0.053 & 1.0$\times$ &  89.2& 89.26&  96.20 & 94.10 & 88.5& 91.45  \\
\midrule
\multirow{4}{*}{FL}
& FFA-LoRA\textsuperscript{*}  & 1.44 & 27.17$\times$ & 88.83  & 88.27 & 94.95 & 91.52 & 86.71 & 89.39 \\
& FedDPA-LoRA\textsuperscript{*} & 2.62 & 49.44$\times$ & 88.99  & 88.43 & 95.50 & 90.74 & 85.73 & 89.47 \\
& FedSA-LoRA\textsuperscript{*} & 1.83  & 10.40$\times$ & 90.18 & 88.88 & 96.00 & 92.13 & 87.48 & 90.43 \\
& FedReFT ($\alpha$=0.5) & 0.053 & 1.0$\times$ &  88.86  & 88.76 & 95.17 & 94.52 & 86.57 & 90.93 \\
& \textbf{FedReFT (TTC)} & \textbf{0.053} & \textbf{1.0$\times$} & \textbf{89.75} & \textbf{89.31} & \textbf{95.75}  & \textbf{94.91}& \textbf{87.15}& \textbf{91.37}\\ 

\bottomrule
\label{tbl:glue_task_ablation}
\end{tabular}
\end{table*}

\begin{table*}[!htbp] 
\setlength{\tabcolsep}{4.5pt}
\centering
\caption{Performance comparison across arithmetic reasoning tasks with the Distinct Task (DT) and Mixed Task (MT) setup using different model sizes. We report results under two mixing strategies: (i) \textbf{adaptive mixing} with Test-Time Computing (TTC), which dynamically balances local and global interventions at inference, and (ii) \textbf{balanced mixing}, where $\alpha=0.5$ denotes the proportion of local contribution and $1-\alpha$ the global contribution.}
\begin{tabular}{c|cccc|cccc}
\toprule
\multicolumn{1}{c|}{} & \multicolumn{4}{c|}{\textbf{Distinct Task (DT)}} & \multicolumn{4}{c}{\textbf{Mixed Task (MT)}} \\
\cmidrule(r){2-5} \cmidrule(l){6-9}
{\textbf{FedReFT ($\alpha=0.5$)}}& \textbf{AQuA} & \textbf{SVAMP} & \textbf{MAWPS} & \textbf{Avg $\uparrow$} & \textbf{AQuA} & \textbf{SVAMP} & \textbf{MAWPS} & \textbf{Avg $\uparrow$} \\
\midrule
LLaMA 7B      &  25.59 & 25.47 & 49.80 & 33.62  &  22.83 & 14.33 & 27.10 & 21.42    \\
LLaMA-2 7B    &  29.53 & 32.45 & 57.30 & 39.76  &  21.65 & 20.39 & 31.50 & 24.51    \\
LLaMA-3 8B    &  34.64 & 48.98 & 73.60 & 52.41  &  31.89 & 48.90 & 70.04 & 50.48    \\
\midrule

\multicolumn{1}{c|}{\textbf{FedReFT (TTC)}} & \multicolumn{4}{c|}{\textbf{Distinct Task (DT)}} & \multicolumn{4}{c}{\textbf{Mixed Task (MT)}} \\
\cmidrule(r){2-5} \cmidrule(l){6-9}
LLaMA 7B      &  26.12 & 26.71 & 49.94 & 34.26  & 20.86 & 15.60 & 28.22 & 22.23   \\
LLaMA-2 7B    &  30.96 & 33.31 & 59.51 & 41.93 & 22.05 & 23.50 & 32.74 & 26.09  \\
LLaMA-3 8B    &  35.36 & 49.41 & 75.25 & 53.34 & 33.45 & 51.28 & 73.51 & 52.75 \\
\bottomrule
\end{tabular}
\label{tbl:math_setup_1_2_resutls_ablation}
\end{table*}
\subsection{Comparison of Aggregation Methods on Different Tasks}
\label{abl:aggregation_methods}
\begin{figure}[htbp]
    \centering
    \includegraphics[width=1\linewidth]{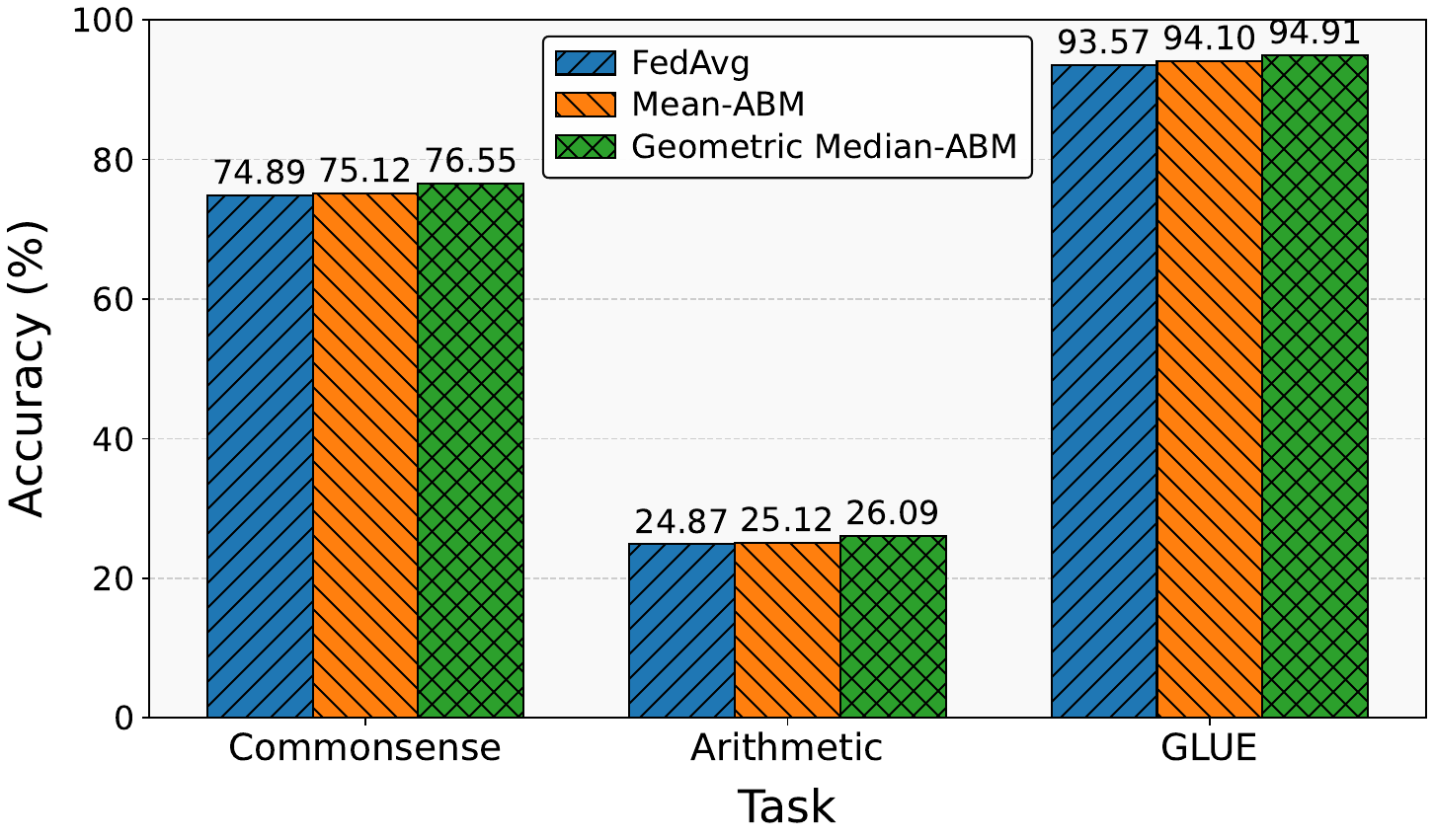}
    \caption{Comparison of aggregation strategies across tasks. Results for FedAvg, Mean-ABM, and Geometric Median-ABM on Commonsense Reasoning, Arithmetic Reasoning, and GLUE show that Geometric Median-ABM consistently outperforms others, demonstrating greater robustness in heterogeneous federated settings.}
    \label{fig:ablation_aggregation_methods_comparison}
\end{figure}

We use training data from the COMMONSENSE170K dataset, split among three clients, and evaluate the models using the SIQA task. Figure~\ref{fig:ablation_aggregation_methods_comparison} shows that Geometric Median ABM aggregation outperforms all other approaches. Similarly, we split the MATH10K dataset among three clients, train each client for only five local epochs, and evaluate the results using the GSM8K evaluation set. Additionally, we use the QNLI GLUE dataset for the natural language understanding (NLU) task.  
\noindent\textbf{Why small gains matter.} We emphasize that even modest accuracy improvements are meaningful in the federated learning setting, particularly under highly heterogeneous task distributions. As shown in Figure~\ref{fig:ablation_aggregation_methods_comparison} the arithmetic mean has time complexity $O(d)$, while the geometric median (Weiszfeld’s algorithm) has $O(T \cdot d)$, with $T$ as iterations and $d$ as parameters. Both methods have memory complexity $O(d)$. Despite the higher computational cost, the geometric median provides greater robustness to heterogeneity and consistently yields higher accuracy in FL settings. 
\begin{table}[!htbp] 
\setlength{\tabcolsep}{0.1pt}
\centering
\caption{Performance comparison on arithmetic reasoning tasks for GSM8K on LLaMA-3 8B model with LoRA rank 8. \textsuperscript{*}Performance results of all baseline methods are taken from \cite{guo2024selective}. \textbf{Trainable Parameter (TP) Efficiency} indicates the efficiency of FedReFT compared to the baselines. $\Delta$\textbf{Acc} shows the accuracy improvement of FedReFT.}
\begin{tabular}{lcccc}
\toprule
\textbf{Method} & \textbf{TP(M)$\downarrow$} & \textbf{TP Effi.$\downarrow$} & \textbf{Acc} & $\Delta$\textbf{Acc}$\uparrow$\\
\midrule
LoReFT  &  4.19 & 1.0$\times$ & 48.33 & +0.55  \\ 
LoRA\textsuperscript{*} &  30.40 & 7.26$\times$ & 46.23  & +2.65\\
\midrule
FedSA-LoRA\textsuperscript{*} &  30.40 & 7.26$\times$ & 46.63 & +2.25 \\
FFA-LoRA\textsuperscript{*} & 15.2 & 3.63$\times$ & 46.32 & +2.56\\

\midrule
 FedReFT (tie $\phi$) & 2.09 & 0.5$\times$ & 47.27 & +1.61\\ 
 FedReFT $\alpha=0.5$ &  4.19 & 1.0$\times$ & 48.39 & +0.49\\

 \textbf{FedReFT(TTC)} &  \textbf{4.19} & \textbf{1.0$\times$} & \textbf{48.88} & \\
\bottomrule
\label{tbl:math_setup_3_baseline_ablation}
\end{tabular}
\end{table}

These improvements hold consistently across three diverse tasks. We chose Geometric Median-ABM not only for its peak accuracy but also for its robustness to outliers and task drift. The benefits are especially pronounced in arithmetic reasoning, where client diversity is highest. While geometric median incurs additional server-side computation, it is applied once per round on low-dimensional sparse parameters, making the overhead negligible. Client-side operations remain lightweight. 
\begin{equation}
f(\mathbf{y}) = \sum_{i=1}^n \|\mathbf{y} - \mathbf{x}_i\|_2.
\end{equation}
\section{Communication Efficiency}
\label{apendix:Communication Efficiency}
As shown in Table~\ref{tab:appendix_trainable_param}, FedReFT is communication and computationally efficient as it uses only a very small percentage of trainable parameters (TP) compared to the total model parameters. For example, in LLaMA-7B and LLaMA-2 7B, only 0.0311\% of the total parameters are trained. In RoBERTa Large, this number is even smaller, at just 0.0138\%. Even for large models like LLaMA-2-13B, the trainable portion remains as low as 0.0503\%. This shows that FedReFT is highly parameter-efficient. Despite using such a small fraction of parameters, FedReFT still achieves strong performance, as discussed in the experimental analysis section \ref{sec:experiments}. This highlights the benefit of using FedReFT in resource-constrained or communication-limited federated learning settings.
\subsection{Intervention Parameter sharing Across Tokens in Same Layer}
In this section, we also conducted some additional experiments to show the robustness of FedReFT in different setups. We experiment with whether to share (tie) the intervention parameters $\phi$ across different input positions within the same layer.
Given the positions \( P = \{1, \ldots, p\} \cup \{n - s + 1, \ldots, n\} \), we define the untied and tied variants \cite{Wu2024}:
\begin{equation}
\begin{aligned}
\mathbf{I}_{\text{untied}} = \left\{ \langle \Phi, \{p\}, l \rangle \mid p \in P,\, l \in L \right\}, \\
\qquad
\mathbf{I}_{\text{tied}} = \left\{ \langle \Phi, P, l \rangle \mid l \in L \right\}.
\end{aligned}
\end{equation}
while FedReFT (tie $\phi$) offers a compelling trade-off between performance and efficiency. These results highlight the scalability and efficiency of our representation-tuning approach.
Appendix Table \ref{tab:fed-sb-performance} and \ref{tbl:math_setup_1_2_resutls_ablation} depicts these. 
\begin{table}[!htbp]
\setlength{\tabcolsep}{1.2pt}
\renewcommand{\arraystretch}{0.99}
\caption{We vary LLaMA model sizes with $C=3$ clients following the Distinct Task (DT) design for the commonsense reasoning task, alongside a centralized LoReFT baseline. As model capacity increases, we observe notable performance gains, with the largest model approaching the accuracy of the centralized setting.
}
\centering
\label{tbl:able_commonsense_DT}
\begin{tabular}{lccccc}
\toprule
\textbf{Method}  & \textbf{BoolQ} & \textbf{PIQA} & \textbf{HellaS.} &  \textbf{Avg $\uparrow$} \\
\midrule

Tiny LLaMA 1B    & 63.92 & 50.36 & 47.21 & 53.83 \\
LLaMA 7B         & 66.64 & 78.34 & 67.92 & 70.97 \\
LLaMA-2 7B       & 69.71 & 75.45 & 78.26 & 74.47 \\
LLaMA-3.2 3B     & 65.93 & 78.37 & 82.36 & 75.55 \\

\bottomrule
\label{tbl:second_setup_commonsense}
\end{tabular}
\end{table}
\subsection{Additional Experimental Validation}
We experiment with different LLaMA model sizes across $C=3$ clients following the Distinct Task (Dt) framework for the commonsense reasoning task in Table \ref{tbl:able_commonsense_DT}, along with a centralized LoReFT baseline for comparison. As the model size grows, we observe steady performance improvements, with the largest variant achieving results close to the centralized model. The first four experiments are conducted under the standalone (centralized) setup, and the following four are performed in the federated learning (FL) environment.
\begin{table}[!htbp]
\setlength{\tabcolsep}{2.7pt}
\centering
\caption{Trainable Intervention Parameters across Models (in Millions) in FedReFT}
\begin{tabular}{lrrr}
\toprule
\textbf{Model} & \textbf{Total P(M)} & \textbf{TP(M)} & \textbf{TP(\%)$\downarrow$} \\
\midrule
LLaMA-1.1B & 1100.05 & 0.72 & 0.0655 \\
LLaMA 7B & 6,738.42 & 2.10 & 0.0311 \\
LLaMA-2 7B& 6,738.42 & 2.10 & 0.0311 \\
LLaMA-3 8B & 8,030.27 & 2.10 & 0.0261 \\
LLaMA-2-13B & 13,015.86 & 6.55 & 0.0503 \\
RoBERTa Large & 355.36 & 0.050 & 0.0138 \\
\bottomrule
\label{tab:appendix_trainable_param}
\end{tabular}
\end{table}
\section{Dataset Description}
\label{appendix:dataset_description}
\subsection{Commonsense Reasoning}
We train and evaluate our models on eight commonsense reasoning datasets spanning different types of open-ended QA tasks, following \cite{Hu21}, we construct all examples. Table \ref{tab:commonsense_examples} shows the dataset samples. 

\begin{itemize}
    \item \textbf{BoolQ} \cite{clark2019boolq}: A yes/no question answering dataset consisting of naturally occurring questions. We remove the associated passages to ensure a fair comparison.
    \item \textbf{PIQA} \cite{bisk2020piqa}: A dataset for physical commonsense reasoning. The model must select the more plausible solution to everyday physical tasks.
    \item \textbf{SIQA} \cite{sap2019socialiqa}: Focuses on social interaction reasoning by asking the model to choose responses based on human intent and consequences.
    \item \textbf{HellaSwag} \cite{zellers2019hellaswag}: Requires choosing the most coherent sentence completion given a context, often involving physical or temporal common sense.
    \item \textbf{WinoGrande} \cite{sakaguchi2021winogrande}: Inspired by the Winograd Schema Challenge \cite{levesque2012winograd}, this dataset contains fill-in-the-blank problems with binary choices requiring commonsense coreference reasoning.
\end{itemize}

We follow the experimental setup in \cite{Hu21} by fine-tuning our models on a combined training corpus referred to as Commonsense170K, which merges all of the above datasets. Evaluation is conducted individually on each dataset’s test split.
\begin{table*}[!htbp]
\centering
\small
\setlength{\tabcolsep}{6pt}
\renewcommand{\arraystretch}{1.25}
\caption{Examples from commonsense reasoning tasks: BoolQ~\cite{clark2019boolq}, PIQA~\cite{bisk2020piqa}, HellaSwag~\cite{zellers2019hellaswag}, and SIQA~\cite{sap2019socialiqa}. Each instruction is followed by the selected answer during evaluation.}
\vspace{3pt}
\begin{tabular}{|p{1.2cm}|p{11.8cm}|p{1.1cm}|}
\hline
\textbf{Dataset} & \textbf{Instruction / Question} & \textbf{Answer} \\
\hline

BoolQ &
\textit{Please answer the following question with true or false:} \newline
Question: Do Iran and Afghanistan speak the same language? &
True \\
\hline

PIQA &
\textit{Please choose the correct solution to the question:} \newline
Question: When boiling butter, when it's ready, you can \newline
Solution1: Pour it onto a plate \newline
Solution2: Pour it into a jar &
Solution2 \\
\hline

HellaSwag &
\textit{Please choose the correct ending to complete the given sentence:} \newline
Removing ice from car: Then, the man writes over the snow covering the window of a car, and a woman wearing winter clothes smiles. then \newline
Ending1: , the man adds wax to the windshield and cuts it. \newline
Ending2: , a person boards a ski lift... \newline
Ending3: , the man puts on a christmas coat... \newline
Ending4: , the man continues removing the snow on his car. &
Ending4 \\
\hline

SIQA &
\textit{Please choose the correct answer to the question:} \newline
Cameron decided to have a barbecue and gathered her friends together. \newline
How would others feel as a result? \newline
Answer1: like attending \newline
Answer2: like staying home \newline
Answer3: a good friend to have &
Answer1 \\
\hline
\end{tabular}
\label{tab:commonsense_examples}
\end{table*}

\begin{table*}[!htbp]
\centering
\small
\setlength{\tabcolsep}{6pt}
\renewcommand{\arraystretch}{1.25}
\caption{Examples from math reasoning tasks: AQuA~\cite{ling2017program}, GSM8K~\cite{cobbe2021training}, SVAMP~\cite{patel2021nlp}, and MAWPS~\cite{koncel2016mawps}. Each instruction is followed by the correct answer derived through step-by-step reasoning.}
\vspace{3pt}
\begin{tabular}{|p{1.2cm}|p{11.8cm}|p{1.2cm}|}
\hline
\textbf{Dataset} & \textbf{Instruction / Question} & \textbf{Answer} \\
\hline

AQuA &
\textit{Solve the following word problem:} \newline
A car is driven in a straight line toward the base of a vertical tower. It takes 10 minutes for the angle of elevation to change from 45° to 60°. After how much more time will the car reach the base of the tower? \newline
Answer Choices: (A) 5($\sqrt{3}$ + 1), (B) 6($\sqrt{3}$ + $\sqrt{2}$), (C) 7($\sqrt{3}$ – 1), (D) 8($\sqrt{3}$ – 2), (E) None of these. &
(A) \\
\hline

GSM8K &
\textit{Solve the following question:} \newline
Janet’s ducks lay 16 eggs per day. She eats 3 eggs and uses 4 for baking. She sells the rest at \$2 per egg. How much money does she make daily? &
\$18 \\
\hline

SVAMP &
\textit{Solve the following arithmetic question:} \newline
Each pack of DVDs costs \$76. A discount of \$25 is applied. What is the final price per pack? &
\$51 \\
\hline

MAWPS &
\textit{Solve the following math word problem:} \newline
Tom has 3 boxes of pencils, each containing 12 pencils. He gives 7 pencils to his friend. How many pencils does Tom have left in total? &
29 \\
\hline
\end{tabular}
\label{tab:math_examples}
\end{table*}

\begin{table*}[!htbp]
\centering
\small
\setlength{\tabcolsep}{6pt}
\renewcommand{\arraystretch}{1.25}
\caption{Examples from GLUE benchmark~\cite{wang2018glue} tasks: MNLI, SST-2, QNLI, and QQP. Each instruction is followed by the corresponding ground truth label.}
\vspace{3pt}
\begin{tabular}{|p{1.2cm}|p{11.8cm}|p{1.2cm}|}
\hline
\textbf{Dataset} & \textbf{Instruction / Question} & \textbf{Answer} \\
\hline

MNLI &
Premise: The dog is running through the field. \newline
Hypothesis: An animal is moving. \newline
Label: entailment &
Entailment \\
\hline

SST-2 &
Sentence: A touching and thought-provoking piece of cinema. \newline
Label: positive &
Positive \\
\hline

QNLI &
Question: What is the capital of France? \newline
Sentence: Paris is the capital and most populous city of France. \newline
Label: entailment &
Entailment \\
\hline

QQP &
Question1: How do I learn to play guitar? \newline
Question2: What is the best way to learn guitar? \newline
Label: duplicate &
Duplicate \\
\hline
\end{tabular}
\label{tab:glue_examples}
\end{table*}

\subsection{Arithmetic Reasoning}
We evaluate arithmetic reasoning using seven benchmark datasets that cover a range of math word problem types. As in \cite{Hu21}, we construct all examples without using golden or retrieved passages. Data samples are shows in Table \ref{tab:math_examples}.

\begin{itemize}
    \item \textbf{AQuA}~\cite{ling2017program}: Presents algebraic word problems in a multiple-choice format.
    \item \textbf{GSM8K}~\cite{cobbe2021training}: A widely used benchmark of grade-school math problems requiring multi-step reasoning.
    \item \textbf{SVAMP}~\cite{patel2021nlp}: A more challenging dataset that tests robustness to paraphrased and structurally altered word problems.
    \item \textbf{MAWPS}~\cite{koncel2016mawps}: A large repository of math word problems aggregated from multiple sources, covering diverse arithmetic and algebraic reasoning tasks expressed in natural language.
    
\end{itemize}

Following \cite{Hu21}, we train our models on a combined training set named \textbf{MATH10K}.
\subsection{Natural Language Understanding}
For NLU, we evaluate on the GLUE benchmark following the evaluation protocol in \cite{Wu2024}. Data samples for shown in Table \ref{tab:glue_examples}.

\begin{itemize}
    \item The validation set is split into two subsets one for in-training evaluation and the other for final testing.
    \item For large datasets (QQP, MNLI, QNLI), 1,000 samples are used for in-training validation.
    \item For smaller datasets, half of the validation set is used during training.
\end{itemize}

\section{Computational Resources}
All experiments are executed on a single NVIDIA A100-SXM4-80GB GPU, except for LLaMA-2 13B, which is run on a GPUH200x8 141GB system to accommodate the computational demands of large-scale federated fine-tuning.

\end{document}